\documentclass[10pt,twocolumn,letterpaper]{article}

\usepackage[pagenumbers]{cvpr} %

\usepackage{amsfonts}
\usepackage{multirow}
\usepackage{cuted}      %
\usepackage{tikz}
\usetikzlibrary{arrows.meta,positioning,fit,backgrounds,calc,shadows.blur}
\usepackage{microtype}

\newcommand{\model}{\mbox{STRING}}
\newcommand{\simname}{\mbox{STRING World}} %
\newcommand{\schema}[1]{\textup{\textsc{#1}}} %
\usepackage{xspace}
\usepackage{colortbl} %
\usepackage{algorithm}
\usepackage{algpseudocode}
\algrenewcommand\alglinenumber[1]{\footnotesize #1}
\renewcommand{\eg}{e.g.,\xspace}

\definecolor{cvhuman}{RGB}{214,120,40}
\definecolor{cvsim}{RGB}{38,110,190}
\definecolor{cvrobot}{RGB}{60,140,90}
\definecolor{cvagent}{RGB}{120,90,190}
\definecolor{cvpanel}{RGB}{244,246,250}

\usepackage{fontawesome5}
\usetikzlibrary{fadings}
\tikzfading[name=swRim, inner color=transparent!100, outer color=transparent!74]%
\definecolor{fgLav}{RGB}{227,232,251}
\definecolor{fgGrey}{RGB}{238,240,243}
\definecolor{fgPeach}{RGB}{251,227,210}
\definecolor{fgOrange}{RGB}{179,77,10}
\definecolor{fgHi}{RGB}{236,107,20}
\definecolor{fgBlue}{RGB}{123,144,225}
\definecolor{fgInk}{RGB}{20,23,28}
\definecolor{fgSub}{RGB}{105,108,115}
\definecolor{fgTab}{RGB}{185,190,198}
\definecolor{fgTeal}{RGB}{52,150,128}
\definecolor{fgBlueD}{RGB}{78,104,204}
\definecolor{fgEdge}{RGB}{160,165,174}
\newcommand{\figfont}{\fontfamily{LinuxBiolinumT-LF}\selectfont}
\newcommand{\figmono}{\fontfamily{SourceCodePro-TLF}\selectfont}

\definecolor{cvprblue}{rgb}{0.21,0.49,0.74}
\usepackage[pagebackref,breaklinks,colorlinks,allcolors=cvprblue]{hyperref}

\def\confName{CVPR}
\def\confYear{2027}

\title{\model{}: A Schema-Thinking Robot Agent That Learns\\ from Embodied Experience in a Multiphysics World}
\author{%
  Zhi Cao$^{*}$ \quad Zesen Zhao$^{*}$ \quad Zixuan Lu$^{*}$ \quad Yining Hong$^{*}$ \quad Peihao Li \quad Yin Yang\\
  Zhuoqing Mao \quad Chenfanfu Jiang \quad Li Fei-Fei \quad Jiajun Wu$^{\dagger}$ \quad Huang Huang$^{\dagger}$\\[3pt]
  {\normalsize $^{*}$Equal contribution \quad $^{\dagger}$Equal advising}
}

\begin{document}
\maketitle
\input{sections/teaser}

\begin{abstract}
Dexterous robots usually learn deformable, contact-rich tasks from teleoperated demonstrations, and the resulting policy works only where the data reached.
Robot agents built on large language models need no such data, but they accumulate skills that overfit to the task they were found on.
We argue that an agent is all you need for dexterous manipulation if it learns as people do, by abstracting physical schemas from embodied experience instead of collecting skills.
We present \model{}, a system in which a schema-based robot agent gains this experience in \simname{}, a multiphysics engine we built from scratch.
\simname{} gives each class of material its own solver, from elastic solids and thin shells to fracturing, granular and liquid matter, strings with audio, and air, and simulates them together with the robot in one scene, covering more kinds of matter than simulators such as MuJoCo and Isaac Sim.
It returns the physical quantities it computes, such as contact reactions and solver state, to the agent as \emph{observables}.
From one human demonstration, the agent rehearses the task in \simname{} and abstracts \emph{physical schemas}, whose roles, relations, phases and force ranges are predicates over the observables.
The schemas and their subgoals carry from simulation to reality, where the robot reproduces the task from a real-time 3D reconstruction of the scene, with no teleoperation data or policy network.
On a bimanual dexterous robot, \model{} reaches 50.0\% average success over ten tasks, from knotting a rope to flying a paper plane with a hair dryer, against 2.5\% for the best general-purpose agent given the same demonstration and \simname{}.
On the four hardest tasks it reaches 43.8\%, against 27.5\% for the best fine-tuned policy.
\end{abstract}

\vspace{-0.6em}
\section{Introduction}
\label{sec:intro}
Teaching a robot hand to knot a rope or fold a shirt today means teaching it a policy, fitted to teleoperated demonstrations~\cite{chi2023diffusionpolicy,black2024pi0}, to human videos converted into robot trajectories~\cite{lum2025crossing,wang2026dexagent}, or to rewards in simulation~\cite{qi2023rotateit,lin2024twisting}.
For deformable, contact-rich objects these data are slow to collect, and the policy rarely works far from where they were recorded.
Robot agents built on large language models skip the data: they plan over skill libraries~\cite{ahn2022saycan}, write policy code~\cite{liang2023code}, and accumulate skills, procedures and task memories as they go~\cite{wang2023voyager,jia2026agp,fang2026wicl}.
But each such skill, program or memory overfits to the task and scene it was found in, and every new task needs a new one.

People learn differently: shown a knot once, a person tries it a few times, partly in their head~\cite{battaglia2013simulation}, and then ties it.
They do not copy the motion; they recognize its structure.
A shoelace, a rope and a drawstring bag are tied by one pattern: a flexible member is tensioned, crossed over itself, held by friction and released.
Cognitive science calls such patterns \emph{schemas}~\cite{bartlett1932remembering,rumelhart1980schemata,mandler1992image}, abstracted from embodied experience and carried to distant domains by \emph{metaphor}~\cite{lakoff1980metaphors,johnson1987body,gentner1983structure}.
We ask whether a robot agent can learn the same way, abstracting schemas rather than collecting skills: \emph{watch once, rehearse,~perform}.

Two things are missing.
First, the agent needs a world in which to gain embodied experience, where cables, cloth, solids, liquids and air respond to the robot and sheets to the air, and where the forces that decide the outcome, such as a thumb's pressure on a deck of cards, can be read.
General-purpose simulators~\cite{todorov2012mujoco,makoviychuk2021isaac,genesis2024} are built around rigid bodies, cover only some of these materials, rarely simulate air, and report contact forces at most, not air impulses or the state of the solve.
Second, the agent needs a form for what it learns that, unlike a skill, outlasts the task and scene it was~found~in.

We present \textbf{\model{}}, a system in which a schema-based robot agent learns from embodied experience in a world we built for it, \textbf{\simname{}} (\cref{fig:teaser}).
\simname{} is a multiphysics engine written from scratch, with a dedicated solver for each class of material, from elastic solids and thin shells to liquids, strings and air.
All solvers share one clock and the robot that drives them, and every object and contact pair returns what the solvers compute to the agent as observables.

The agent \emph{watches} one human demonstration and takes it as the reference.
It \emph{rehearses} the task in \simname{}, exploring robot actions while it fits the object's material, and abstracts what it experiences into \emph{physical schemas}: roles, relations, phases and force ranges, written as predicates over the observables.
The phases split the task into subgoals, for the rope cross, thread and tighten.
It then \emph{performs} on a bimanual robot: from a real-time 3D reconstruction of the scene, it invokes the same schemas and follows the reference trajectory rehearsed in \simname{}, with each predicate turned into a goal the cameras can check.
Because it practices in \simname{} and keeps schemas, not trajectories, skills or network weights, the agent carries what it learns to reality and to later tasks (\cref{fig:schema}) without teleoperation data or a policy network: an agent is all you need for dexterous manipulation.
Our contributions~are:
\begingroup\setlength{\topsep}{2pt}\setlength{\partopsep}{0pt}
\begin{itemize}\setlength{\itemsep}{1pt}\setlength{\parsep}{0pt}
\item \textbf{\simname{}}, a multiphysics engine built from scratch that gives each class of material its own solver, FEM for elastic solids and thin shells, MPM for fracturing, granular and liquid matter, and dedicated solvers for strings, with the audio they produce, and for air, runs them in one scene driven by the robot, couples thin shells and air in one energy-consistent step, and reports the contact reactions, air impulses and solver state it computes to the agent. To our knowledge it is the first engine to cover all these kinds of matter; MuJoCo, Isaac Sim, Genesis and Newton each cover only some (\cref{tab:simcompare}).
\item \textbf{Schemas instead of skills}, a paradigm in which a robot agent abstracts physical schemas, a pattern of human thinking, from embodied experience in \simname{} rather than accumulating task-fitted skills, and reproduces in reality what it rehearsed from one human demonstration.
\item \textbf{Real-robot results}: one agent, acting as the policy, completes ten dexterous manipulation tasks that span cables, cloth, paper, liquids, granular and fracturing solids and air, with long multi-step bimanual motion, at 50.0\% average success without teleoperation data or a policy network, against 2.5\% for the best agent baseline; on the four hardest tasks it reaches 43.8\%, against 27.5\% for the best learned policy.
\end{itemize}\endgroup

\section{Related Work}
\label{sec:related}

\textbf{Simulation for manipulation.}
Robot learning runs mostly on articulated-body engines such as MuJoCo~\cite{todorov2012mujoco} and Isaac~\cite{makoviychuk2021isaac}; MuJoCo~4.0 adds implicit deformables and penetration-free cloth~\cite{mujoco4}, and Genesis~\cite{genesis2024} and Newton~\cite{newton2025} bring MPM, PBD and VBD solvers to the GPU.
Deformable benchmarks each build on one solver family~\cite{lin2020softgym,macklin2014unified,huang2021plasticinelab,xian2023fluidlab,wang2024thinshelllab,liu2024softmac,heiden2021disect, geng2025roboverse}.
These environments serve a learner and cover a few materials each; \simname{} is built for an agent, simulates all of these kinds of matter together with the robot in one scene, and returns the reaction of every contact pair as an observable.
On the solver side we build on MPM with plasticity and damage~\cite{jiang2016mpm,wolper2019cdmpm,wolper2020anisompm}, on barrier-free contact~\cite{zheng2025barrierfree} in place of the log barrier of IPC~\cite{li2020ipc,li2021cipc,huang2024gipc}, and on Robin coupling~\cite{badia2008robin,fernandez2013robin}, rather than lagged coupling~\cite{guendelman2005coupling}, of shells to a flow-map air solver~\cite{wang2025cirrus}.

\textbf{Deformable manipulation from demonstrations.}
Rope, cloth, dough and cards are usually handled by policies trained on many interactions or teleoperated demonstrations~\cite{nair2017rope,grannen2020untangling,avigal2022speedfolding,ha2021flingbot,shi2023robocook,chen2026dexholdem,zhao2026axis}, and dexterous sim-to-real work likewise transfers a trained policy~\cite{qi2023rotateit,lin2024twisting, zhu2026tucocuratingsimulationdemonstrations, yan2026mattersqualityawarevlapretraining}.
Human videos can be retargeted to robot hands~\cite{qin2022dexmv,shaw2023videodex,li2024okami,lepert2025phantom, li2025rose, chen2026large} or lifted into simulation through hand and object reconstruction~\cite{pavlakos2024hamer,yu2025dynhamr,torne2024rialto,lum2025crossing,jiang2025phystwin,memmel2024asid}, and ORION~\cite{zhu2026orion} and DexAgent~\cite{wang2026dexagent} turn a single video into an object-graph plan or an agentic human-to-simulation-to-robot pipeline.
In all of these, what reaches the robot is a trajectory, a plan over object poses or a trained policy; in \model{} it is the schemas and subgoals found in rehearsal, and no policy is trained.

\textbf{Agents for manipulation.}
Language models plan over skill libraries~\cite{ahn2022saycan,huang2022inner}, write policy code~\cite{liang2023code}, compose value maps or keypoint constraints~\cite{huang2023voxposer,huang2024rekep}, accumulate skills~\cite{wang2023voyager}, explain failures~\cite{liu2023reflect,liu2026recova, AlphaOne25}, identify physical parameters~\cite{cherian2026llmphy} and generate tasks and rewards~\cite{wang2023robogen,wang2023gensim,ma2023eureka}.
Closest to ours, Agent as Policy drives a real robot through programs that it revises and reuses across trials~\cite{jia2026agp}, and the concurrent World In-Context Learning rehearses a task in simulation from one demonstration and deploys the resulting task memory~\cite{fang2026wicl}.
These agents keep programs, steps, skills or object roles, each fitted to the task it was found on; \model{} keeps physical schemas abstracted from embodied experience, with predicates over forces and phases, and carries them to the robot as goals it~can~see.

\textbf{Schemas and analogy.}
Schema theory describes understanding as the reuse of relational patterns~\cite{bartlett1932remembering,rumelhart1980schemata}; image schemas arise from bodily experience~\cite{johnson1987body,mandler1992image,hedblom2020image}, metaphor and structure mapping extend them to new domains~\cite{lakoff1980metaphors,gentner1983structure,falkenhainer1989sme}, and people are thought to understand physical scenes by mental simulation~\cite{battaglia2013simulation}.
Schemas have been learned for games~\cite{kansky2017schema}, large language models solve many analogy problems without examples~\cite{webb2023emergent}, and image schemas have been proposed as a grounding for agent reasoning~\cite{olivier2025grounding}.
We make them concrete on a robot, as predicates over simulated physical quantities that are checked through their visible consequences.

\section{Method}
\label{sec:method}

\model{} has two parts (\cref{fig:teaser}; notation in Appendix~\ref{sec:supp-notation}).
\simname{}, the simulator, gives each class of material its own solver, runs the solvers together with the robot in one scene and couples thin shells with air (\cref{sec:solid,sec:gas}), and returns what they compute to the agent as observables (\cref{sec:signals}).
The agent follows \emph{watch once, rehearse, perform}: it rehearses the human demonstration in \simname{} while reconstructing the object (\cref{sec:rehearse}), abstracts what it experiences into physical schemas instead of skills (\cref{sec:schema}), splits the task into subgoals (\cref{sec:subgoal}), and performs it on the robot (\cref{sec:perform}).

\subsection{Solid and Liquid Solvers}
\label{sec:solid}
Each body uses the solver that suits its material: finite elements (FEM) for elastic solids and thin shells, the material point method (MPM) for matter that flows, compacts or fractures, a fine-stepped 1D model for strings and an incompressible solver for air (\cref{sec:gas}), all on one clock, each substepping at its own rate within a control frame.
The robot, two YAM arms with Sharpa hands, is position controlled and enters every solver as a prescribed moving boundary (meshes in FEM, spheres in MPM, solid faces in air); what objects exert on it is read out as contact reactions, the simulated touch.

\textbf{FEM.}
Ropes are tetrahedral hyperelastic solids, and cloth, cards, paper and film are codimensional shells with a membrane energy per triangle and a hinge energy per interior edge~\cite{grinspun2003discrete}, whose rest angle yields plastically with hardening so paper keeps a crease.
Implicit Euler minimizes the incremental potential
\begin{equation}
E(\mathbf x) = \tfrac12(\mathbf x-\tilde{\mathbf x})^{\!\top}\mathbf M(\mathbf x-\tilde{\mathbf x}) + h^2 \big(U(\mathbf x)+U_f(\mathbf x)\big)
\label{eq:ip}
\end{equation}
over vertex positions $\mathbf x$, with lumped mass $\mathbf M$, step $h$, ballistic target $\tilde{\mathbf x} = \mathbf x^n + h\mathbf v^n + h^2\mathbf M^{-1}\mathbf f_{\rm ext}$, elastic energy $U$ and the IPC friction potential $U_f$~\cite{li2020ipc}, a smoothed Coulomb potential of each pair's slip with normal force and tangent frame lagged by one step.

\input{sections/figure_solvers}
\textbf{Contact.}
Every point--triangle and edge--edge distance $d_i$, self-contact included, must stay above the pair's thickness along the step; instead of a log barrier~\cite{li2020ipc} we keep this a constraint~\cite{zheng2025barrierfree}, linearized at the last intersection-free iterate $\mathbf x^{[k]}$,
\begin{equation}
\begin{split}
\hat{\mathbf x}^{[k+1]} &= \arg\min_{\hat{\mathbf x}} E(\hat{\mathbf x})\ \ \text{s.t.}\ \ c_i(\hat{\mathbf x})\ge 0,\\
c_i(\hat{\mathbf x}) &= d_i + \nabla d_i^{\top}(\hat{\mathbf x}-\mathbf x^{[k]}) - \delta_i,
\end{split}
\label{eq:sub}
\end{equation}
with $\delta_i$ the thickness plus a small gap.
Projected Newton with PCG minimizes an augmented Lagrangian over the active set $I_C$ with slack $s_i\ge0$, after which slack and multiplier update in closed form,
\begin{align}
\mathcal L &= E(\hat{\mathbf x}) + \textstyle\sum_{i\in I_C}\gamma_i\big[\tfrac{\mu}{2}(c_i-s_i)^2 - \lambda_i(c_i-s_i)\big],\label{eq:al}\\
s_i&\leftarrow\max\!\big(0,\,c_i-\lambda_i/\mu\big),\quad \lambda_i\leftarrow\lambda_i-\mu(c_i-s_i).\label{eq:update}
\end{align}
The penalty $\mu$, a tenth of the largest diagonal entry of $\mathbf M+h^2\nabla^2U$, is fixed within the step, which keeps the conditioning bounded into contact.
The iterate moves toward $\hat{\mathbf x}$ only as far as CCD certifies,
\begin{equation}
\mathbf x^{[k+1]} = (1-\alpha^{[k+1]})\,\mathbf x^{[k]} + \alpha^{[k+1]}\,\hat{\mathbf x}^{[k+1]},
\label{eq:blend}
\end{equation}
with $\alpha^{[k+1]}\in[0,1]$ the largest collision-free fraction.
CCD hits join $I_C$, separated pairs decay as $\gamma_i\leftarrow\Gamma\gamma_i$ until pruned, and the step ends when the weight $\omega$ that $\mathbf x^n$ and the first iterates carry in $\mathbf x^{[k]}$ falls below $\epsilon$ (Appendix~\ref{sec:supp-contact}), leaving $\mathbf f_{c,i} = h^{-2}\lambda_i\nabla d_i(\mathbf x)$ as the explicit normal force on pair $i$.

\textbf{Strings.}
Strings vibrate at kilohertz, too fast for the implicit step, so each is a chain of nodes under the exact prestressed axial force, with Euler--Bernoulli bending and frequency-dependent loss~\cite{bilbao2009numerical}, stepped by symplectic Euler over a thousand times per FEM step and touched through a penalty with Coulomb friction; its anchor forces drive a soundboard and cavity that radiate to a microphone (Appendix~\ref{sec:supp-strings}).

\textbf{MPM.}
For matter that tears, compacts or splashes, MPM~\cite{jiang2016mpm} particles carry mass $m_p$, rest volume $V_p^0$, position $\mathbf x_p$, velocity $\mathbf v_p$, affine velocity $\mathbf C_p$ and deformation gradient $\mathbf F_p=\mathbf F^E_p\mathbf F^P_p$, split into elastic and plastic parts, and a background grid rebuilt every step solves the momentum equation.
A step is explicit under a CFL limit, runs on the GPU in Warp~\cite{macklin2022warp}, and has four stages.
\par\noindent(1)~\emph{Advection} moves each particle, $\mathbf x_p\leftarrow\mathbf x_p+h\mathbf v_p$, and deforms it, $\mathbf F^E_p\leftarrow\mathcal Z\big((\mathbf I+h\mathbf C_p)\mathbf F^E_p\big)$, where the material's return mapping $\mathcal Z$ projects $\mathbf F^E_p$ onto its elastic region and moves the excess into $\mathbf F^P_p$.
\par\noindent(2)~\emph{P2G} transfers mass and affine momentum to grid nodes $i$ at $\mathbf x_i$, spacing $\Delta x$, with quadratic B-spline weights $w_{ip}$ (APIC~\cite{jiang2015apic}),
\begin{equation}
m_i\mathbf v_i=\textstyle\sum_pw_{ip}m_p\big(\mathbf v_p+\mathbf C_p(\mathbf x_i-\mathbf x_p)\big).
\label{eq:p2g}
\end{equation}
(3)~\emph{Grid update and projection} adds stress and gravity and projects onto the collider constraints with $\Pi_i$,
\begin{equation}
\mathbf v_i^{+} = \Pi_i\Big(\mathbf v_i - \tfrac{h}{m_i}\textstyle\sum_p V_p^0\,\mathbf P(\mathbf F^E_p)\,\mathbf F^{E\top}_p\nabla w_{ip} + h\,\mathbf g\Big).
\label{eq:grid}
\end{equation}
(4)~\emph{G2P} gathers the new velocity and its affine part,
\begin{equation}
\mathbf v_p=\textstyle\sum_iw_{ip}\mathbf v_i^{+},\qquad \mathbf C_p=\tfrac{4}{\Delta x^2}\textstyle\sum_iw_{ip}\mathbf v_i^{+}(\mathbf x_i-\mathbf x_p)^{\top}.
\label{eq:g2p}
\end{equation}
A material is thus a stress $\mathbf P$ and a return mapping $\mathcal Z$.
\emph{Fracturing solids} (bread) follow CD-MPM~\cite{wolper2019cdmpm}: a phase-field damage $d_p\in[0,1]$, grown by tensile strain energy past a threshold, degrades shear and tensile stress by $(1-d_p)^2$, and a damage-softened von Mises yield adds dents.
\emph{Granular solids} (snow) follow~\cite{stomakhin2013snow}: $\mathcal Z$ clamps the singular values of $\mathbf F^E_p$ to $[1-\theta_c,1+\theta_s]$, the moduli harden by $e^{\xi(1-J^P_p)}$ with $J^P_p=\det\mathbf F^P_p$, and $\theta_s$ rises with compaction so packed snow coheres.
\emph{Liquids} (water) use Tait pressure $p=\frac{K}{\gamma}(J_p^{-\gamma}-1)$ without tension, $J_p$ advanced by the velocity divergence, and FLIP blended into G2P.
Nodes inside a robot link (fitted spheres) or a tool or container (signed distance) lose their approaching normal velocity and keep a Coulomb-limited tangential one, or take the collider's velocity under a sticky grip (Appendix~\ref{sec:supp-mpm}).

\subsection{Gas Solver for Thin Shells in Air}
\input{sections/figure_schema}
\label{sec:gas}
A paper sheet weighs about as much as the air it pushes, an added-mass ratio at which partitioned schemes diverge~\cite{causin2005added}, so we solve shell and air in one step.
The incompressible, inviscid air lives on an adaptive GPU octree with cell pressures and MAC face velocities $u$, refined around the sheet and advected with particle flow maps, as in Cirrus~\cite{wang2025cirrus}.
A face whose cell-center segment crosses a sheet triangle is a \emph{wall} that moves with the sheet's normal speed while the sheet slides freely along it.
Let $\mathbf v$ be the shell vertex velocities, $N$ the map to the normal speed at each wall, $\tau_f=e_f\!\cdot n_f$ the tilt between face axis $e_f$ and triangle normal $n_f$, and $C=D_{\mathcal W}\,\mathrm{diag}(\tau)$ the wall part of the divergence.
After advection to $u^*$, projecting air of density $\rho$ with walls moving at speeds $\nu$ is
\begin{equation}
\begin{aligned}
e_a(\nu)=\min_{u_{\mathcal O}}\ &\tfrac12\textstyle\sum_{f\in\mathcal O}\rho V_f(u_f-u^*_f)^2\\
\text{s.t.}\ \ &D_{\mathcal O}u_{\mathcal O}+C\nu+D_{\mathcal S}u_{\mathcal S}=0,
\end{aligned}
\label{eq:ea}
\end{equation}
a convex quadratic $e_a=\tfrac12 r^{\top}L^{-1}r$ with $r=D_{\mathcal O}u^*+C\nu+D_{\mathcal S}u_{\mathcal S}$ over open faces $\mathcal O$ and prescribed solid faces $\mathcal S$, closed-wall Laplacian $L$, and gradient $\nabla e_a=-C^{\top}q$ for the pressure impulse $q=hp$.
With $e_s(\mathbf v)=E(\mathbf x^n+h\mathbf v)/h^2$, one coupled step solves
\begin{equation}
\min_{\mathbf v}\ e_s(\mathbf v)+e_a(N\mathbf v),
\label{eq:coupled}
\end{equation}
whose optimality conditions are the shell's implicit Euler step under the air's pressure impulse and the air's projection with walls at the sheet's normal speed (Appendix~\ref{sec:supp-air}); the impulse is the exact gradient of $e_a$, so the exchange is energy-consistent.

We split \cref{eq:coupled} with a copy $\nu$ of $N\mathbf v$, multiplier $J$ and per-wall penalty $\beta_f$, and run consensus ADMM~\cite{boyd2011admm}, whose sweeps solve the shell (\cref{sec:solid}) with a quadratic wall term, the air with a Robin condition at the walls, and the dual,
\begin{gather}
\mathbf v^{k}=\arg\min_{\mathbf v}\ e_s(\mathbf v)+\textstyle\sum_f\tfrac{\beta_f}{2}\big((N\mathbf v)_f-\hat\nu^k_f\big)^2,\label{eq:shellsweep}\\
(L+C\beta^{-1}C^{\top})\,q^{k+1}=-\bar r^k,\label{eq:robin}\\
\nu^{k+1}=\bar\nu^k+\beta^{-1}C^{\top}q^{k+1},\quad J^{k+1}=C^{\top}q^{k+1},\label{eq:dual}
\end{gather}
with $\hat\nu^k=\nu^k+J^k/\beta$, $\bar\nu^k=N\mathbf v^k-J^k/\beta$ and $\bar r^k=D_{\mathcal O}u^*+C\bar\nu^k+D_{\mathcal S}u_{\mathcal S}$.
The multiplier $J$ is the air's impulse on the sheet, and a Robin wall is an open face with fluid-column mass $\beta_f/\tau_f^2$, so the multigrid applies unchanged.
We set $\beta_f$ to the sheet mass the wall stands for, $m_f=\rho_sA_f|\tau_f|$.
The first sweep solves the shell under the previous step's impulse without the wall term, and after the last sweep the air is projected again with closed walls at the committed sheet velocity, with one constant flux correction per trapped air pocket (Appendix~\ref{sec:supp-air}).

\Cref{fig:solvers} shows two of them at work, air coupled to the shell of a paper plane and MPM water poured into a beaker, each beside one step of its solver.

\subsection{Physical Observables for the Agent}
\label{sec:signals}
Beyond rendered frames and a reward, \simname{} returns per step, object and contact pair the physical quantities its solves compute, its \emph{observables}: \emph{forces} $\partial E/\partial\mathbf x$ ($\lambda_i$, $\mathbf f_{c,i}$, window $\lambda_i/\mu$, friction and slip of each typed pair, air impulse $J_f$), \emph{state} (motion, kinetic energy and reversals, FEM strain and hinge curvature, MPM damage $d_p$ and $J^P_p$) and the \emph{solver} record (active-set changes, CCD hits, $\alpha$, $\omega$, Newton and PCG iterations, air sweeps, residuals and leak).
Failure codes, task metrics and checkpoints complete the interface, reached through a declarative scene specification~\cite{mcp2024}.

\subsection{Rehearsal from One Demonstration}
\label{sec:rehearse}
Rehearsal is where the agent gains embodied experience: it acts in \simname{} and reads the observables.
From one human demonstration recorded by three calibrated cameras, the agent reconstructs MANO hands~\cite{romero2017mano,pavlakos2024hamer,yu2025dynhamr} and the object, and loads the object into \simname{} as MPM particles, a shell mesh or a cable.
It then retargets the hand motion to the robot and explores actions while fitting the object's material parameters to the recorded motion.
Rehearsal ends when the robot reaches every subgoal of the demonstration in simulation (\cref{sec:subgoal}); Appendix~\ref{sec:supp-real} details the pipeline.

\subsection{Physical Schemas Instead of Skills}
\label{sec:schema}
Agents for manipulation usually accumulate skills, programs or task memories~\cite{ahn2022saycan,liang2023code,wang2023voyager,jia2026agp,fang2026wicl}, each of which overfits to the task and scene it was found in.
\model{} keeps physical schemas instead, the relational patterns in which people are thought to reason about physical tasks~\cite{johnson1987body,mandler1992image}; skills remain only as executable primitives.

\textbf{Schema.}
A schema is a tuple $\sigma=\langle R,\Phi,P,G,F,\Theta\rangle$ of roles $R$ (\eg member, anchor, support), relations $\Phi$ over the roles, predicates $P$ that ground each relation in the observables, a phase graph $G$ (\eg approach $\to$ contact $\to$ load $\to$ release), failure signatures $F$, and parameters $\Theta$ such as force bands and timing.
\schema{Tension}, for example, binds a member between two anchors: it is active when both anchors carry opposing reactions $\lambda_i$ with no slip, $u_t\approx0$, and it fails when $\lambda_i$ drops at one anchor.

\textbf{From observables to tasks.}
The agent reasons over four levels (\cref{fig:schema}): physical quantities from \simname{}, the events that threshold rules detect in them, the schemas that relate the events, and tasks, each an \emph{assembly} $\mathcal A$ of the schemas active together with their role bindings and phase order.
The agent starts from a short text on how people structure physical problems, which contains no rule of the form ``for task X, call skill Y'', and stores each schema it forms as a JSON record (\cref{fig:schema}, right) with its roles, guidance and the observables that are its evidence (Appendix~\ref{sec:supp-schema}).

\textbf{Agent loop.}
A main agent with reviewer, diagnostician and verifier sub-agents forms the \emph{harness}, which runs five steps in rehearsal.
(1)~It recognizes the assembly $\mathcal A_T$ of the task in the replayed demonstration.
(2)~It retrieves the closest stored assembly by the structural score~\cite{gentner1983structure}
\begin{equation}
S(\mathcal A,\mathcal A') = \max_{\pi}\ \frac{|\{\phi\in\Phi_{\mathcal A} : \pi(\phi)\in\Phi_{\mathcal A'}\}|}{|\Phi_{\mathcal A}|}
\label{eq:sim}
\end{equation}
over role bijections $\pi$, which counts shared relations and ignores object attributes, so a rope matches a shoelace rather than a rod.
(3)~It binds the roles to the scene and initializes $\Theta$ from the match.
(4)~It simulates and diagnoses: when a rollout fails, the predicates and failure signatures locate the relation and contact pair that broke, and the agent repairs that parameter and restarts from the last checkpoint.
(5)~It stores $\mathcal A_T$ with its parameters and failure signatures.
A force band or recovery found on one task is thus available to every later task that contains the same relation.

\input{sections/figure_multimodal}

\subsection{Subgoals}
\label{sec:subgoal}
The phase graphs of the active schemas split a task into subgoals, such as crossing the rope, threading the loop and tightening the knot.
A subgoal is defined by the predicates that must hold when it is reached and by a checkpoint to return to, not by a pose or a stretch of trajectory, so a failure, in simulation or on the robot, restarts from the last subgoal that still holds.

\subsection{Performing on the Robot}
\label{sec:perform}
Rehearsal yields an agentic policy rather than network weights: the schemas with their subgoals and parameters, and the reference trajectory that reached every subgoal in \simname{}.
The same harness runs it on the robot, one subgoal at a time.
It reconstructs the scene in 3D in real time from three RGB-D cameras, locates the subgoal's target and moves the plan onto it, fits the hand to its measured pose, and executes the reference trajectory re-solved on the measured scene rather than replayed.
It then reconstructs the scene again and checks the subgoal's visible goal; a failed check undoes the stage and retries it with a small change.
In one rope run, registration moved the plan by 5.8\,cm, and two fits of the hand reduced its landing error from 5.9 to 0.5\,cm before contact (Appendix~\ref{sec:supp-real}).
Before transfer, the agent also renders the rollout in a scene matched to the lab, following DexAgent~\cite{wang2026dexagent}, to test the camera-based checks (\cref{fig:real}).

\textbf{Schemas without their simulated quantities.}
Contact multipliers, slip or air impulses have no sensor on the robot, and the schema does not need them there.
During rehearsal the agent compiles each predicate into a goal and a consequence that the reconstruction can see.
The grasp goal for a cloth corner is the mesh vertex where both finger pads first press with more than 0.5\,N in \simname{}, and ``the fist holds both strands'' becomes the share of rope voxels left on the table after the lift, 94\% after a first close in one run and 0\% after a retry 1\,cm lower.
Quantities with no visible consequence are used only in \simname{}, where they select and verify the reference trajectory.

\section{Experiments}
\label{sec:exp}

\begin{table}[t]
\caption{\textbf{Simulator comparison} by the kinds of matter each one simulates with its own solver, two-way shell--air coupling, and the observables reported to the agent: {\color{fgTeal}\checkmark}\ native, $\circ$ partial, -- none, following each simulator's documentation as of November 2026.}
\label{tab:simcompare}
\centering\footnotesize
\setlength{\tabcolsep}{2.6pt}
\setlength{\aboverulesep}{0pt}\setlength{\belowrulesep}{0pt}
\renewcommand{\arraystretch}{1.14}
\newcommand{\hd}[1]{\begin{tabular}[b]{@{}c@{}}#1\end{tabular}}
\newcommand{\ic}[1]{{\color{fgSub}\scriptsize#1}}
\newcommand{\y}{{\color{fgTeal}$\checkmark$}}
\newcommand{\p}{{\color{fgSub}$\circ$}}
\newcommand{\n}{{\color{fgTab}--}}
\newcolumntype{G}{>{\columncolor{fgGrey!70}}c}
\resizebox{\columnwidth}{!}{%
\begin{tabular}{@{}lGcGcGc@{\hspace{6pt}}Gc@{}}
\toprule
 & \multicolumn{6}{c}{Matter, one solver each} & \multicolumn{2}{c}{Coupling and observables} \\
\cmidrule(lr){2-7}\cmidrule(l){8-9}
 & \ic{\faTshirt} & \ic{\faHammer} & \ic{\faSnowflake} & \ic{\faTint} & \ic{\faGuitar} & \ic{\faWind} & \ic{\faArrowsAltH} & \ic{\faStethoscope} \\[-1pt]
 & \hd{Solids and\\shells$^{a}$} & \hd{Fracturing\\solids} & Granular & Liquid & \hd{Strings\\with audio} & Air & \hd{Two-way\\shell--air} & \hd{Forces and\\solver state} \\
\midrule
MuJoCo~4.0 & \p & \n & \n & \n & \n & \p & \n & \p \\
\rowcolor{fgGrey} Isaac Sim & \p & \n & \y & \y & \n & \n & \n & \p \\
Genesis & \y & \p & \y & \y & \n & \p & \n & \p \\
\rowcolor{fgGrey} Newton & \p & \p & \y & \n & \n & \n & \n & \p \\
\rowcolor{fgLav} \textbf{\simname{}} & \y & \y & \y & \y & \y & \y & \y & \y \\
\bottomrule
\end{tabular}}
\\[2pt]\raggedright\scriptsize $^{a}$\,\checkmark\ with intersection-free contact. $\circ$\,Solids and shells: FEM without guaranteed intersection-free contact (MuJoCo: guaranteed for 2D cloth only). Fracture: MPM material separates without a damage model. Air: a drag model for bodies (MuJoCo), a stable-fluid smoke solver with no documented coupling to shells (Genesis). Forces: contact forces only, without air impulses or the state of the contact solve.
\end{table}

\begin{table}[t]
\caption{\textbf{Cost per step} of each sub-solver on one NVIDIA RTX~5090. \#$p$/\#$f$: vertices or particles and triangles; $\Delta x$: grid cell; $h$: time step; $t$: wall-clock time per step.}
\label{tab:perf}
\centering\footnotesize
\setlength{\tabcolsep}{1.85pt}
\setlength{\aboverulesep}{0pt}\setlength{\belowrulesep}{0pt}
\renewcommand{\arraystretch}{1.28}
\newcommand{\sci}[2]{$#1{\times}10^{#2}$}
\resizebox{\columnwidth}{!}{%
\begin{tabular}{@{}llrrrrr@{}}
\toprule
Task & Sub-solver & \#DOF & \#$p$/\#$f$ & $\Delta x$ (mm) & $h$ (s) & $t$ (ms) \\
\midrule
\rowcolor{fgGrey} Pour water & MPM & 483k & 161k/-- & 1.25 & \sci{2.1}{-5} & 1.25 \\
 & Strings$^{a}$ & 21.9k & 7{,}401/-- & -- & \sci{2.0}{-3} & 38.5 \\
\multirow{-2}{*}{Harp} & Audio$^{b}$ & 5{,}549 & 1{,}547/2{,}880 & -- & \sci{2.1}{-5} & 0.023 \\
\rowcolor{fgGrey} Snowball & MPM & 1.71M & 570k/-- & 5.21 & \sci{4.9}{-5} & 1.53 \\
Bread tear & MPM & 2.71M & 903k/-- & 2.60 & \sci{3.5}{-5} & 7.61 \\
\rowcolor{fgGrey} Card shuffle & FEM shells & 17.0k & 5{,}670/9{,}072 & -- & \sci{2.0}{-3} & 355 \\
 & FEM shell & 8{,}889 & 2{,}963/5{,}718 & -- & & 33.2$^{d}$ \\
\multirow{-2}{*}{Paper plane} & Air & 1.10M & -- & 3.91$^{c}$ & \multirow{-2}{*}{\sci{1.28}{-4}} & 67.8$^{d}$ \\
\bottomrule
\end{tabular}}
\\[3pt]\raggedright\scriptsize $^{a}$\,each string advanced by fine steps of \sci{6.5}{-7} to \sci{1.9}{-6}\,s inside the step. $^{b}$\,offline, on the CPU. $^{c}$\,finest octree level, coarsest 62.5\,mm. $^{d}$\,one coupled step takes 102.7\,ms.
\end{table}

We evaluate the two parts of \model{} in turn: \simname{} as a simulator, by the matter it covers and what a step costs (\cref{sec:exp-sim}), and the whole system on a real robot, where an agent that watched one demonstration and rehearsed in \simname{} must reproduce the task (\cref{sec:exp-real}).
An ablation then separates what the harness, \simname{} and the initial state contribute (\cref{sec:exp-ablation}).

\subsection{\simname{}: Coverage and Performance}
\label{sec:exp-sim}
\textbf{Coverage.}
\Cref{fig:multimodal,fig:simroll} show twelve simulated tasks, none with a task-specific solver. They include paper that keeps its creases through plastic hinges, a pipette that draws liquid by air pressure, and three demanding regimes over time: a paper plane blown into flight by the air jet of a hair dryer, snow that keeps the shape pressed into it, and a riffle of 52 cards under dense self-contact.
\Cref{tab:simcompare} compares \simname{} with four general-purpose simulators.
Only Genesis guarantees intersection-free contact, through an optional coupler; fracture appears elsewhere only as MPM material that separates without a damage model; and no other simulator has strings that produce audio or couples a thin shell with the air around it in both directions.
All four report contact forces, but not air impulses or the state of the contact solve.

\textbf{Performance.}
\Cref{tab:perf} gives the cost of one step of each sub-solver on the GPU, averaged over the first five minutes of each task.
Explicit MPM is the cheapest, 1.3 to 7.6\,ms for 0.16 to 0.9 million particles, or 119 to 373 million particle updates per second.
The implicit solvers take larger steps at a higher cost: 38.5\,ms for the harp's strings, 355\,ms for the 52-card riffle and 102.7\,ms for the coupled step of the paper plane.
No scene runs in real time, from about 19\,s of wall-clock time per simulated second for the harp to about 13\,min for the paper plane.

\subsection{Real-Robot Transfer from One Demonstration}
\label{sec:exp-real}
\input{sections/figure_real}
\begin{table}[t]
\caption{\textbf{Real-robot success rate} (\%) of agents given one human demonstration per task and no teleoperation data, over 20 trials per task; the reasoning effort of each baseline is in brackets. Time is the average execution time of \model{}. Best result per task in bold.}
\label{tab:real}
\centering\footnotesize
\newcommand{\best}[1]{\textbf{#1}}
\newcommand{\mh}[1]{\begin{tabular}[b]{@{}c@{}}#1\end{tabular}}
\setlength{\tabcolsep}{3pt}
\setlength{\aboverulesep}{0pt}\setlength{\belowrulesep}{0pt}
\renewcommand{\arraystretch}{1.18}
\resizebox{\columnwidth}{!}{%
\begin{tabular}{@{}lcccc@{\hspace{8pt}}r@{}}
\toprule
Task & \mh{GPT-6 Astra\\{}[Max]} & \mh{Opus 5.5\\{}[High]} & \mh{Fable 5.1\\{}[High]} & \mh{\textbf{\model{}}\\(ours)} & \mh{Time\\(ours)} \\
\midrule
Knot the rope             & 0  & 0  & 0  & \best{65} & 13.6 min \\
\rowcolor{fgGrey} Tie the shoelace          & 0  & 0  & 0  & \best{25} & 15.8 min \\
Fold the T-shirt          & 10 & 15 & 10 & \best{60} & 9.1 min \\
\rowcolor{fgGrey} Shuffle the cards         & 0  & 0  & 0  & \best{80} & 52 s \\
Fold a paper plane        & 0  & 0  & 0  & \best{30} & 10.3 min \\
\rowcolor{fgGrey} Fold a paper box          & 0  & 0  & 0  & \best{40} & 3.7 min \\
Use the pipette           & 5  & 0  & 0  & \best{40} & 8.1 min \\
\rowcolor{fgGrey} Cut and serve the bread   & 0  & 0  & 0  & \best{65} & 4.0 min \\
Make a snowball           & 10 & 5  & 0  & \best{65} & 3.2 min \\
\rowcolor{fgGrey} Fly a plane by hair dryer & 0  & 0  & 0  & \best{30} & 5.9 min \\
\midrule
Mean                      & 2.5 & 2.0 & 1.0 & \best{50.0} & 7.5 min\\
\bottomrule
\end{tabular}}
\end{table}
\begin{table}[t]
\caption{\textbf{Comparison with learned policies} on the four hardest tasks, success rate (\%). The policies are fine-tuned per task on 10 teleoperated and 90 human demonstrations; \model{} uses one human demonstration and no teleoperation data.}
\label{tab:policy}
\centering\footnotesize
\newcommand{\best}[1]{\textbf{#1}}
\setlength{\tabcolsep}{4pt}
\setlength{\aboverulesep}{0pt}\setlength{\belowrulesep}{0pt}
\renewcommand{\arraystretch}{1.18}
\resizebox{\columnwidth}{!}{%
\begin{tabular}{@{}lccc@{\hspace{8pt}}c@{}}
\toprule
 & FACT~\cite{anon2026fact} & T-Rex~\cite{niu2026trex} & $\pi_{0.5}$~\cite{black2025pi05} & \textbf{\model{}} \\
\midrule
Tie the shoelace          & 20 & 0  & 0 & \best{25} \\
\rowcolor{fgGrey} Use the pipette           & \best{40} & 0 & 0 & \best{40} \\
Shuffle the cards         & 30 & 10 & 0 & \best{80} \\
\rowcolor{fgGrey} Fly a plane by hair dryer & 20 & 10 & 0 & \best{30} \\
\bottomrule
\end{tabular}}
\end{table}

\textbf{Setup.}
The robot has two YAM arms with five-finger Sharpa hands, a head camera and two wrist cameras.
Each of ten tasks (\cref{tab:real}) is demonstrated once by a person and recorded by three calibrated cameras; this is the only task-specific input.
A trial succeeds only if it reaches every subgoal: for example, the rope must be crossed, threaded and tightened into a knot, and no card may fly out of the riffle.
The agent is a Codex thread running the harness of \cref{sec:schema}.
The agent baselines, GPT-6 Astra at maximum reasoning effort and Claude Opus~5.5 and Claude Fable~5.1 at high effort, receive the same demonstration, robot interface and \simname{}, but no harness.
On the four hardest tasks we also compare with learned policies, FACT~\cite{anon2026fact}, T-Rex~\cite{niu2026trex} and $\pi_{0.5}$~\cite{black2025pi05}, each fine-tuned per task on 10 teleoperated demonstrations and 90 human demonstrations recorded with a tactile glove.

\textbf{Results.}
\model{} succeeds in 25--80\% of trials across the ten tasks, 50.0\% on average (\cref{tab:real}); \cref{fig:real} follows four tasks from demonstration to rollout and Appendix~\ref{sec:supp-results} all ten.
Without the harness, GPT-6 Astra succeeds only in 2 of 20 T-shirt folds, 1 pipetting trial and 2 snowballs, Opus~5.5 in 3 T-shirt folds and 1 snowball, and Fable~5.1 in 2 T-shirt folds, for 2.5\%, 2.0\% and 1.0\% on average.
Execution takes from 52\,s for the riffle to 15.8\,min for the shoelace, 7.5\,min on average (\cref{tab:real}).

\textbf{Data efficiency.}
On the four hardest tasks (\cref{tab:policy}), FACT reaches 20--40\%, T-Rex at most 10\%, and $\pi_{0.5}$ fails every trial, each after fine-tuning on 100 demonstrations per task, 10 of them teleoperated.
\model{} matches FACT on pipetting and exceeds every policy on the other tasks, 43.8\% on average against 27.5\% for FACT, from a single human demonstration, with no teleoperation data and no training.

\textbf{Failure analysis.}
Success is highest for the riffle, with two subgoals (80\%), and lowest where the last subgoal needs a precise force on a thin part: the final pull of the shoelace (25\%), and the third fold and the launch of the paper plane (30\% each).
Failures concentrate at one step per task (per-subgoal rates in Appendix~\ref{sec:supp-subgoal}), where a small force on a thin or soft part decides the outcome.

\subsection{Ablation Study}
\label{sec:exp-ablation}
We remove one component of \model{} at a time and run the same agent on three simulated tasks; the qualitative results are in Appendix~\ref{sec:supp-ablation} (\cref{fig:baseline}).
The \emph{bare agent} has neither the harness nor \simname{}; \emph{+ harness} adds the harness on MuJoCo; \emph{full \model{}} runs the harness on \simname{}.

\textbf{Harness.}
Without the harness the agent sets wrong cloth parameters and leaves the T-shirt unfolded.
With it, on the same kind of simulator, the agent folds the sleeves and once across the body: the schemas turn a stalled attempt into a structured one (\cref{fig:baseline}, top).

\textbf{\simname{}.}
On MuJoCo the structured attempt still ends with crumpled cloth, and only on \simname{}'s FEM cloth, with idealized hand control, does it end in a flat, squared garment.
The bare agent serves two solid halves of bread and the harness on MuJoCo a loaf of blocks that splits in two, while \simname{}'s damage-aware MPM tears the loaf and sheds crumbs.
For the hair dryer, the bare agent and MuJoCo move a rigid plane under an assumed jet or a drag approximation, and only \simname{} deforms the sheet in air solved together with it (\cref{sec:gas}).

\textbf{Initial state.}
We start the rope-knotting policy on the real robot from two arm and rope configurations (\cref{fig:pose}).
Both runs pass through the same subgoals and complete the knot within 4\,s of each other, because the roles of its schemas, such as \schema{Support Redistribution}, are bound to the rope as it lies rather than to a recorded trajectory.

\section{Conclusion}
\label{sec:conclusion}
We presented \model{}, in which a robot agent abstracts physical schemas, rather than task-fitted skills, from embodied experience in \simname{}, a multiphysics engine built from scratch.
From one human demonstration per task, with no teleoperation data or policy network, it completes ten real-robot tasks at 50.0\% average success, against 2.5\% for the best agent baseline.

\textbf{Limitations.}
No scene of \simname{} runs in real time, and a robot rollout takes up to 16 minutes.
Schema transfer across tasks is shown on one pair of tasks, and how far a schema carries to new objects is not yet measured.
Sim-to-real still loses accuracy: the hardest tasks succeed in 25--30\% of trials, with schemas checked only by camera.

\iftoggle{cvprfinal}{%
\section*{Acknowledgments}
The authors thank Sharpa and Xdof for the donation of the dexterous hands and robot arms used in the experiments.
}{}

\section*{LLM Usage Statement}
Large language models are part of the method: the agent and the agent baselines are LLM-based coding agents, named in \cref{sec:exp-real}, and they write the simulation and robot programs that produce our results.
We also used an LLM assistant to edit the text, typeset figures and tables from our own data, and check references.
All content was reviewed by the authors, who take full responsibility for it.

{
    \small
    \bibliographystyle{ieeenat_fullname}
    \bibliography{references}
}

\clearpage
\appendix

\noindent\textbf{Overview.}
The appendix follows the order of the paper.
Appendix~\ref{sec:supp-notation} collects the notation of \cref{sec:method}, and Appendices~\ref{sec:supp-contact} to~\ref{sec:supp-air} complete the description of \simname{} (\cref{sec:solid,sec:gas,sec:signals}), one sub-solver each.
Appendices~\ref{sec:supp-schema} and~\ref{sec:supp-real} detail the agent: its schema records and harness (\cref{sec:schema}), and how it watches the demonstration and performs on the robot (\cref{sec:rehearse,sec:perform}).
Appendices~\ref{sec:supp-simroll} to~\ref{sec:supp-ablation} extend the experiments: the rollouts of all simulated tasks (\cref{sec:exp-sim}), the real-robot results by subgoal (\cref{sec:exp-real}), and the frames behind the ablation (\cref{sec:exp-ablation}).

\section{Notation and Overview of the Sub-Solvers}
\label{sec:supp-notation}
\Cref{tab:notation} lists the symbols of \cref{sec:method} and of the appendices, and \cref{tab:solvers} summarizes the four sub-solvers that Appendices~\ref{sec:supp-contact} to~\ref{sec:supp-air} describe.
Each of these appendices states the continuous problem that its sub-solver discretizes and adds the details that \cref{sec:solid,sec:gas} omit; equations given in the main text are cited rather than repeated, except where an algorithm listing needs them.

\begin{table*}[t]
\caption{\textbf{Notation}, grouped by topic. A symbol that recurs in another group has the meaning given in that group.}
\label{tab:notation}
\centering\footnotesize
\setlength{\tabcolsep}{4pt}
\setlength{\aboverulesep}{0pt}\setlength{\belowrulesep}{0pt}
\renewcommand{\arraystretch}{1.13}
\newcommand{\nhead}[1]{\rowcolor{fgGrey}\multicolumn{2}{@{}l@{}}{\rule{0pt}{8pt}\hspace{2pt}\textbf{#1}}\\}
\begin{minipage}[t]{0.488\textwidth}\vspace{0pt}
\begin{tabular}{@{}l>{\raggedright\arraybackslash}p{5.55cm}@{}}
\toprule
\nhead{Time stepping, FEM and contact}
$\mathbf x,\mathbf x^n,\mathbf v^n$ & positions; positions and velocities at the start of a step \\
$h,\mathbf M$ & time step, lumped mass matrix \\
$\tilde{\mathbf x}$ & ballistic target $\mathbf x^n+h\mathbf v^n+h^2\mathbf M^{-1}\mathbf f_{\rm ext}$ \\
$E,U,U_f$ & incremental potential, elastic energy, friction potential \\
$d_i,\nabla d_i$ & distance of primitive pair $i$ and its gradient \\
$\delta_i=\delta_i^0+\hat d$ & contact offset: pair thickness plus contact gap \\
$c_i,s_i$ & linearized distance constraint and its slack \\
$\lambda_i,\mu$ & multiplier ($h^2$ times the normal force $\bar\lambda_i$), penalty \\
$\mathbf f_{c,i}$ & normal force on pair $i$, $h^{-2}\lambda_i\nabla d_i$ \\
$\mathbf u_i,\mu_{f,i},\epsilon_v$ & tangential slip over the step (magnitude $u_t$), friction coefficient, slip-speed threshold \\
$I_C,\gamma_i,\Gamma$ & active set, pair weight, decay factor \\
$\alpha^{[k]},\eta$ & collision-free step fraction, CCD safety factor \\
$\omega_j,\omega,\epsilon,K_{\min}$ & weight of iterate $j$, termination weight, its tolerance, number of iterates in the stopping sum \\
$\theta,\bar\theta,\theta_p,k_b$ & dihedral, rest and plastic angles of a hinge, bending stiffness \\
\nhead{Strings and audio}
$\mu_s,T_0,EA,EI$ & linear density, tension, axial and bending stiffness \\
$\sigma_0,\sigma_1$ & loss coefficients \\
$\mathbf F_j,\mathbf y_j$ & force of string $j$ on its soundboard anchor, anchor point \\
$w,m_b,\mathcal D$ & board displacement, areal mass, plate operator \\
$p_c,V,S_h,\xi_h$ & cavity pressure and volume, sound-hole area, displacement of its air plug \\
$q_n,\omega_n,\zeta_n,\phi_n$ & amplitude, frequency, damping ratio and shape of mode $n$ \\
$p(t)$ & audio signal, the pressure at the microphone \\
\bottomrule
\end{tabular}
\end{minipage}\hfill
\begin{minipage}[t]{0.488\textwidth}\vspace{0pt}
\begin{tabular}{@{}l>{\raggedright\arraybackslash}p{5.75cm}@{}}
\toprule
\nhead{MPM}
$m_p,V_p^0,\mathbf x_p,\mathbf v_p$ & particle mass, rest volume, position, velocity \\
$w_{ip},\Delta x$ & B-spline weight of node $i$ at particle $p$, grid spacing \\
$\mathbf C_p,\zeta,\chi$ & affine velocity, rotational blend, FLIP share \\
$\mathbf F_p=\mathbf F^E_p\mathbf F^P_p$ & deformation gradient, elastic and plastic parts \\
$\mathbf P,\mathcal Z,\Pi_i$ & first Piola--Kirchhoff stress, return mapping, collider projection at node $i$ \\
$d_p,H_p$ & damage and its history field \\
$J_p,J^P_p$ & volume ratio, plastic volume ratio $\det\mathbf F^P_p$ \\
$\theta_c,\theta_s,\xi$ & snow critical compression and stretch, hardening \\
$K,\gamma$ & bulk modulus and exponent of the Tait pressure \\
\nhead{Air and shell--air coupling}
$u,u^*$ & face velocities (MAC) and their values after advection \\
$q=hp,\tilde q$ & pressure impulse, Pa$\cdot$s; that of the closed-wall projection \\
$A_f,\delta_f,V_f$ & face area, distance between its cell centers, control volume $A_f\delta_f$ \\
$D,L$ & divergence, closed-wall Laplacian $D_{\mathcal O}(\rho V)^{-1}D_{\mathcal O}^{\top}$ \\
$\mathcal O,\mathcal W,\mathcal S$ & open faces, walls, solid faces \\
$\tau_f,n_f,b_f$ & wall tilt $e_f\!\cdot n_f$, triangle normal, barycentric weights \\
$N,C$ & vertex velocities $\to$ normal speed at the walls; wall part $D_{\mathcal W}\,\mathrm{diag}(\tau)$ of the divergence \\
$e_s,e_a$ & shell and air energies of the coupled step \\
$\nu,J,\beta_f$ & wall-speed copy, multiplier (the air's impulse), penalty \\
$\kappa_f$ & Robin face coefficient $A_f^2\tau_f^2/\beta_f$ \\
$m_f,\rho_s$ & sheet mass per wall $\rho_sA_f|\tau_f|$, sheet areal density \\
$\varrho$ & contraction factor of one sweep \\
$\mathcal P,Q_{\mathcal P}$ & pocket of trapped air, its net wall flux \\
\nhead{Schemas}
$\sigma$ & schema $\langle R,\Phi,P,G,F,\Theta\rangle$: roles, relations, predicates, phase graph, failure signatures, parameters \\
$\mathcal A_T,S(\mathcal A,\mathcal A')$ & assembly of task $T$, structural score of two assemblies \\
\bottomrule
\end{tabular}
\end{minipage}
\end{table*}

\begin{table*}[t]
\caption{\textbf{The four sub-solvers} at a glance. Time steps are those of \cref{tab:perf}; a shell coupled to air steps with the air.}
\label{tab:solvers}
\centering\footnotesize
\setlength{\tabcolsep}{4pt}
\setlength{\aboverulesep}{0pt}\setlength{\belowrulesep}{0pt}
\renewcommand{\arraystretch}{1.16}
\newcommand{\rr}{\raggedright\arraybackslash}
\begin{tabular}{@{}>{\rr}p{1.45cm}>{\rr}p{2.6cm}>{\rr}p{2.9cm}>{\rr}p{3.4cm}>{\rr}p{2.6cm}>{\rr}p{2.7cm}@{}}
\toprule
Sub-solver & Matter & State & Time stepping & The robot enters as & Reported to the agent \\
\midrule
FEM (App.~\ref{sec:supp-contact}) & elastic solids; thin shells: cloth, cards, paper, film & vertex positions of tetrahedra and of triangles with hinges & implicit Euler, augmented-Lagrangian contact certified by CCD; $h=2$\,ms & prescribed vertices of the link meshes & $\lambda_i$, $\mathbf f_{c,i}$, slip, strain, hinge curvature, solver record \\
\rowcolor{fgGrey} Strings (App.~\ref{sec:supp-strings}) & strings under tension, with soundboard and cavity & nodes of a chain; modes of the board & symplectic Euler in fine steps of 0.65 to 1.9\,$\mu$s; modal filters at the audio rate & finger triangles, by penalty with Coulomb friction & anchor forces, audio signal $p(t)$ \\
MPM (App.~\ref{sec:supp-mpm}) & fracturing solids, granular solids, liquids & particles $\mathbf x_p,\mathbf v_p,\mathbf C_p,\mathbf F_p$ on a background grid & explicit under a CFL limit; $h=21$ to $49\,\mu$s & spheres fitted to the links; signed-distance walls for tools & damage $d_p$, plastic volume ratio $J^P_p$ \\
\rowcolor{fgGrey} Air (App.~\ref{sec:supp-air}) & air around thin shells & pressure impulses and face velocities on an octree & flow-map advection, then a projection coupled to the shell by ADMM; $h=0.128$\,ms & solid faces with a prescribed velocity & impulse $J_f$, sweeps, residuals, leak \\
\bottomrule
\end{tabular}
\end{table*}

\section{FEM Sub-Solver}
\label{sec:supp-contact}
\textbf{Problem.}
The FEM bodies obey
\begin{equation}
\mathbf M\ddot{\mathbf x} = -\nabla U(\mathbf x)+\mathbf f_{\rm ext}+\textstyle\sum_i \bar\lambda_i\nabla d_i(\mathbf x)+\mathbf f_{f},
\label{eq:fem-ode}
\end{equation}
where $\mathbf f_{\rm ext}$ is the external force, $\mathbf f_f$ is the friction force, and each primitive pair $i$ carries a normal force $\bar\lambda_i$ that satisfies the Signorini conditions
\begin{equation}
d_i(\mathbf x)-\delta_i\ge0,\qquad \bar\lambda_i\ge0,\qquad \bar\lambda_i\,\big(d_i(\mathbf x)-\delta_i\big)=0
\label{eq:signorini}
\end{equation}
at all times.
In the offset $\delta_i=\delta_i^0+\hat d$, with $\hat d$ the contact gap, a shell contributes half its thickness to the pair thickness $\delta_i^0$, and the surface of a volumetric body contributes nothing.
The friction force of a pair opposes its tangential slip with a magnitude of at most $\mu_{f,i}\bar\lambda_i$, a bound that is attained while the pair slides.
Implicit Euler, with friction lagged by one step, turns a step from $\mathbf x^n$ into the constrained minimization of the incremental potential of \cref{eq:ip},
\begin{equation}
\mathbf x^{n+1}=\arg\min_{\mathbf x}\,E(\mathbf x)\ \ \text{s.t.}\ \ d_i(\mathbf x)\ge\delta_i\ \ \forall i,
\label{eq:fem-step}
\end{equation}
over the states that can be reached from $\mathbf x^n$ along a path on which every $d_i$ stays above $\delta_i^0$.
Its Karush--Kuhn--Tucker (KKT) conditions, $\nabla E(\mathbf x^{n+1})=\sum_i\lambda_i\nabla d_i(\mathbf x^{n+1})$ with $\lambda_i\ge0$ and $\lambda_i(d_i-\delta_i)=0$, are the discrete form of \cref{eq:fem-ode,eq:signorini}, with $\lambda_i=h^2\bar\lambda_i$.

\begin{algorithm}[t]
\caption{FEM step with barrier-free contact.}
\label{alg:fem}
\small
\begin{algorithmic}[1]
\State $\tilde{\mathbf x}\gets\mathbf x^n+h\mathbf v^n+h^2\mathbf M^{-1}\mathbf f_{\rm ext}$, \ $\mathbf x^{[0]}\gets\mathbf x^n$
\State Run the broad phase; keep the surviving pairs with their $\lambda_i,\gamma_i$
\State Build the friction terms of \cref{eq:friction} from $\lambda_i^n$
\State $\mu\gets0.1\max_j(\mathbf M+h^2\nabla^2U)_{jj}$
\For{$k=0,1,\dots$}
\State Form $c_i$ at $\mathbf x^{[k]}$ for $i\in I_C$ \Comment{\cref{eq:sub}}
\State $\hat{\mathbf x}\gets\arg\min\mathcal L$ by projected Newton \Comment{\cref{eq:al}}
\State Update $s_i$ and $\lambda_i$ \Comment{\cref{eq:update}}
\State $\alpha\gets\min(1,\eta\,\mathrm{TOI})$ by CCD from $\mathbf x^{[k]}$ to $\hat{\mathbf x}$
\State Add the earliest new impact of each vertex to $I_C$
\State $\mathbf x^{[k+1]}\gets(1-\alpha)\mathbf x^{[k]}+\alpha\hat{\mathbf x}$ \Comment{\cref{eq:blend}}
\State $\gamma_i\gets1$ for pairs in contact, $\gamma_i\gets\Gamma\gamma_i$ for separated pairs
\State Drop the pairs with $\gamma_i<0.01$
\If{$\omega<\epsilon$}
\State \textbf{break}
\EndIf
\EndFor
\State $\mathbf x^{n+1}\gets\mathbf x^{[k+1]}$, \ $\mathbf v^{n+1}\gets(\mathbf x^{n+1}-\mathbf x^n)/h$
\State Export $\lambda_i$, $d_i$, $I_C$ and the iteration counts
\end{algorithmic}
\end{algorithm}

\textbf{Energies.}
A tetrahedral body has the elastic energy $U=\sum_eV_e\Psi(\mathbf F_e)$ over its elements $e$, with rest volume $V_e$ and deformation gradient $\mathbf F_e$.
With Lam\'e parameters $\mu_L,\lambda_L$ and $J_e=\det\mathbf F_e$, the energy density is either stable Neo-Hookean, $\Psi=\frac{\mu_L}{2}(\|\mathbf F_e\|^2-3)+\frac{\hat\lambda}{2}(J_e-\hat\alpha)^2$ with $\hat\lambda=\lambda_L+\mu_L$ and $\hat\alpha=1+\mu_L/\hat\lambda$, or fixed corotated, $\Psi=\mu_L\sum_k(\sigma_k-1)^2+\frac{\lambda_L}{2}(J_e-1)^2$ in the singular values $\sigma_k$ of $\mathbf F_e$.
A shell of thickness $t$ has the membrane energy $tA_e\Psi(\mathbf F_e)$ on the $3\times2$ deformation gradient of each triangle of rest area $A_e$, with St.~Venant--Kirchhoff (StVK) as the default density and the two-dimensional corotated and Neo-Hookean models as alternatives.
Its bending energy is $k_b\frac{|\bar e|}{\bar h}(\theta-\bar\theta-\theta_p)^2$ per interior edge $\bar e$~\cite{grinspun2003discrete}, with rest length $|\bar e|$ and $\bar h=(A_1+A_2)/(3|\bar e|)$ from the rest areas $A_1$ and $A_2$ of the two adjacent triangles.
After each step, a return mapping with linear hardening updates the plastic angle $\theta_p$ wherever the elastic angle exceeds a yield angle set by the yield strain of the sheet; this strain separates paper, which keeps most of a fold, from film, which springs back.

\textbf{Friction.}
For each pair in contact at the end of the previous step, $\lambda_i^n>0$, the tangent frame of the contact plane is built at $\mathbf x^n$ and frozen for the step, and $\mathbf u_i(\mathbf x)\in\mathbb R^2$ is the relative displacement of the two closest points of the pair over the step, expressed in this frame.
The friction potential of \cref{eq:ip} is then
\begin{equation}
h^2U_f(\mathbf x)=\textstyle\sum_i\mu_{f,i}\,\lambda_i^n\,f_0\big(\|\mathbf u_i(\mathbf x)\|\big),
\label{eq:friction}
\end{equation}
with $f_0(y)=y$ for $y\ge\epsilon_vh$ and, for $y<\epsilon_vh$, the cubic $-\frac{y^3}{3\epsilon_v^2h^2}+\frac{y^2}{\epsilon_vh}+\frac{\epsilon_vh}{3}$ that continues it smoothly~\cite{li2020ipc}.
The resulting force is a Coulomb force of magnitude $\mu_{f,i}\lambda_i^n/h^2$ against the slip once a pair slides faster than $\epsilon_v$; below that speed the force rises smoothly from zero with the slip, which approximates static friction.

\textbf{Algorithm.}
Algorithm~\ref{alg:fem} lists one step of the barrier-free contact solver~\cite{zheng2025barrierfree}.
The feasible set of \cref{eq:fem-step} is not convex, and the path condition is not a constraint on $\mathbf x^{n+1}$ alone, which is why the solver linearizes the distances at certified states (\cref{eq:sub}) and advances only along segments checked by continuous collision detection (CCD, \cref{eq:blend}).
At a fixed point of the iteration, where neither the iterate nor the multipliers change, the linearization is exact and the multipliers satisfy the KKT conditions of \cref{eq:fem-step}.
Eliminating the slack turns each contact term of \cref{eq:al} into the one-sided penalty $\gamma_i\frac{\mu}{2}\max(0,\lambda_i/\mu-c_i)^2$, up to a constant, so a pair exerts a force only inside the window $c_i<\lambda_i/\mu$.
Line~7 minimizes the result by Newton's method with a line search, using the system matrix $[\nabla^2E]^++\mu\sum_i\gamma_i\nabla d_i\nabla d_i^{\top}$ over the pairs inside their windows, where $[\nabla^2E]^+$ is the Hessian of $E$ with its element and friction blocks projected onto the positive semidefinite cone.
The penalty $\mu$ of line~4 stays fixed for the whole step; Zheng \etal~\cite{zheng2025barrierfree} report that the safeguard of their solver, which doubles $\mu$ when the collision-free fraction $\alpha$ stalls, never triggers.
In line~9, $\mathrm{TOI}$ is the earliest time of impact along the segment from $\mathbf x^{[k]}$ to $\hat{\mathbf x}$, as a fraction of the segment, and $\eta<1$ is a safety factor; a segment without an impact gives $\alpha=1$.
Line~10 admits a new pair only if it is the earliest impact of one of its vertices, so a fast primitive that sweeps through several layers does not add all of them at once.
Unrolling \cref{eq:blend} gives $\mathbf x^{[k]}=\omega_0\mathbf x^n+\sum_{j=1}^k\omega_j\hat{\mathbf x}^{[j]}$ with $\omega_j=\alpha^{[j]}\prod_{l>j}(1-\alpha^{[l]})$ and $\alpha^{[0]}=1$, and the termination weight of line~14 is $\omega=\sum_{j<K_{\min}}\omega_j$, the share of $\mathbf x^n$ and of the first $K_{\min}-1$ iterates; stopping at $\omega<\epsilon$ instead of at the fixed point keeps the step first-order accurate in $h$~\cite{zheng2025barrierfree}.
The defaults are $\epsilon=10^{-3}$ and $\Gamma=0.9$.
The displacement of the prescribed link vertices over the step enters $c_i$ and the CCD, so free vertices are pushed out of the way of an oncoming link within the same step.

\section{Strings and Audio}
\label{sec:supp-strings}
\textbf{String problem.}
A string with linear density $\mu_s$, speaking length $L$, axial stiffness $EA$ and bending stiffness $EI$, tuned to the tension $T_0$, is a curve $\mathbf x(s,t)$, parameterized by the arc length $s$ of its tuned rest state, that is fixed at the soundboard anchor and at the neck pin and obeys
\begin{equation}
\begin{aligned}
\mu_s\partial_t^2\mathbf x={}&\partial_s(T\,\hat{\mathbf t})-EI\,\partial_s^4\mathbf x\\
&-2\mu_s\sigma_0\partial_t\mathbf x+2\mu_s\sigma_1\partial_s^2\partial_t\mathbf x+\mu_s\mathbf g+\mathbf f_c,
\end{aligned}
\label{eq:string}
\end{equation}
where $\hat{\mathbf t}$ is the unit tangent, $T=EA\big((1+T_0/EA)|\partial_s\mathbf x|-1\big)$ is the axial force of a string prestretched to $T_0$, $\mathbf g$ is gravity, and $\mathbf f_c$ is the finger contact force: a normal force that keeps the string out of the fingers and a tangential force bounded by Coulomb friction.
The tension carries the transverse waves at the speed $\sqrt{T_0/\mu_s}$, and $EA$ carries the longitudinal waves and the tension modulation of a large pluck.
Bending makes the partials inharmonic, with the $l$-th partial at $l\bar f\sqrt{1+Bl^2}$, where $\bar f$ is the fundamental without bending and $B=\pi^2EI/(T_0L^2)$; a mode of wavenumber $k$ decays at the rate $\sigma_0+\sigma_1k^2$~\cite{bilbao2009numerical}.
The string model outputs the force $\mathbf F_j(t)$ that string $j$ exerts on its soundboard anchor $\mathbf y_j$.

\textbf{Audio problem.}
The soundboard is an orthotropic Kirchhoff plate, clamped at its edges, with areal mass $m_b$ and normal displacement $w(\mathbf y,t)$, positive out of the body.
It is driven by the normal components $F_j$ of the anchor forces and by the pressure $p_c$ of the air in the body, a cavity of volume $V$ that opens through sound holes of area $S_h$, whose air plug has mass $m_h$ and displacement $\xi_h$:
\begin{equation}
\begin{gathered}
m_b\,\ddot w+\mathcal Dw=\textstyle\sum_jF_j(t)\,\delta(\mathbf y-\mathbf y_j)+p_c,\qquad m_h\ddot\xi_h=S_hp_c,\\
p_c=-\tfrac{\rho c^2}{V}\Big(\textstyle\int w\,dA+S_h\xi_h\Big),
\end{gathered}
\label{eq:board}
\end{equation}
with $\mathcal D$ the orthotropic plate operator, $\delta$ the Dirac delta, $\rho$ the density of air and $c$ the speed of sound.
The audio signal $p(t)$ is the pressure at a microphone, the sum of the radiation of the board, treated as a baffled plane, and of the sound holes, treated as a monopole,
\begin{equation}
p(t)=\frac{\rho}{2\pi}\int\frac{\ddot w(\mathbf y,t-r/c)}{r}\,dA+\frac{\rho S_h}{4\pi}\frac{\ddot\xi_h(t-r_h/c)}{r_h},
\label{eq:rayleigh}
\end{equation}
with $r$ and $r_h$ the distances from a point of the board and from the sound holes to the microphone.
The coupling is one-way, from the strings to the board, because the end loss of the strings, calibrated to measured decays, already accounts for the energy that the board absorbs.

\textbf{String discretization.}
The nodes of a string are a few millimeters apart, and bending uses a five-point fourth difference.
The fine step (\cref{tab:perf}) is half the longitudinal CFL limit, so one FEM step contains one thousand to three thousand fine steps.
The finger contact is evaluated every few fine steps against the finger triangles that a broad phase finds once per FEM step, and the anchor forces are recorded at the audio sampling rate.

\textbf{Audio discretization.}
The modes of the board are computed with Morley finite elements.
Together with the plug, they turn \cref{eq:board} into one modal problem, which is coupled through the compliance of the cavity and solved once.
Each coupled mode $n$, with amplitude $q_n$, frequency $\omega_n$ and shape $\phi_n$ on the board, is assigned a damping ratio $\zeta_n$ and is then a damped oscillator,
\begin{equation}
\ddot q_n+2\zeta_n\omega_n\dot q_n+\omega_n^2q_n=\textstyle\sum_j\phi_n(\mathbf y_j)\,F_j(t),
\label{eq:modes}
\end{equation}
which is discretized by the bilinear transform at the audio sampling rate, and \cref{eq:rayleigh} becomes a sum over the elements.
The mass and moduli of the board and the compliance of the cavity are calibrated to the measured body modes of the~instrument.

\section{MPM Sub-Solver}
\label{sec:supp-mpm}
\textbf{Problem.}
An MPM body is a continuum with density $\rho$, velocity $\mathbf v$ and deformation gradient $\mathbf F=\mathbf F^E\mathbf F^P$.
With $J=\det\mathbf F$ and the material derivative $D/Dt$, it obeys
\begin{gather}
\frac{D\rho}{Dt}=-\rho\,\nabla\!\cdot\mathbf v,\qquad \frac{D\mathbf F}{Dt}=(\nabla\mathbf v)\,\mathbf F,\nonumber\\
\rho\frac{D\mathbf v}{Dt}=\nabla\!\cdot\!\Big(\tfrac1J\,\mathbf P(\mathbf F^E)\,\mathbf F^{E\top}\Big)+\rho\,\mathbf g,
\label{eq:mpm-pde}
\end{gather}
together with a material law: the stress $\mathbf P$ derives from a hyperelastic energy of the elastic part, and a plastic flow keeps $\mathbf F^E$ inside the elastic region of the material.
MPM discretizes the weak form of \cref{eq:mpm-pde}: the particles are quadrature points that carry the state, the grid nodes carry the B-spline basis in which the momentum equation is solved, and the plastic flow is applied per particle as the return mapping $\mathcal Z$ of the trial $\mathbf F^E_p$.

\textbf{Transfers.}
A step runs the four stages of \cref{sec:solid}, with material-dependent changes to the two transfers (\cref{tab:mpm}).
In P2G, the affine velocity can be blended toward its rotational part, $\mathbf C_p\leftarrow(1-\zeta)\mathbf C_p+\tfrac{\zeta}{2}(\mathbf C_p-\mathbf C_p^{\top})$, which damps the elastic transients that contact excites.
Dropping the affine term from \cref{eq:p2g} altogether gives the particle-in-cell (PIC) transfer; the deformation gradient is then still advanced by the velocity gradient gathered in G2P.
For liquids, G2P blends in a share $\chi$ of the fluid-implicit-particle (FLIP) update, $\mathbf v_p\leftarrow\chi\big(\mathbf v_p+\sum_iw_{ip}(\mathbf v_i^+-\mathbf v_i)\big)+(1-\chi)\sum_iw_{ip}\mathbf v_i^+$, because APIC alone damps a sloshing pool within a few~\mbox{periods}.

\begin{table}[t]
\caption{\textbf{MPM materials}, each defined by a stress, a return mapping and a transfer; the degradation is $g(d)=(1-k)(1-d)^2+k$ with a small residual $k$.}
\label{tab:mpm}
\centering\footnotesize
\setlength{\tabcolsep}{3pt}
\setlength{\aboverulesep}{0pt}\setlength{\belowrulesep}{0pt}
\renewcommand{\arraystretch}{1.14}
\newcommand{\rr}{\raggedright\arraybackslash}
\begin{tabular}{@{}>{\rr}p{1.15cm}>{\rr}p{2.2cm}>{\rr}p{2.25cm}>{\rr}p{2.0cm}@{}}
\toprule
 & Fracturing solid (bread) & Granular solid (snow) & Liquid (water) \\
\midrule
Stress & Neo-Hookean, split into tension and compression, degraded by $g(d_p)$ & fixed corotated, moduli scaled by $e^{\xi(1-J^P_p)}$ up to a cap & Tait pressure, $\gamma=7$, zero for $J_p>1$ \\
\rowcolor{fgGrey} Return mapping & von Mises, yield stress scaled by $(1-d_p)^2$ & singular values clamped to $[1-\theta_c,1+\theta_s]$ & none, $\mathbf F_p=J_p^{1/3}\mathbf I$ \\
Extra state & damage $d_p$, history $H_p$ & $J^P_p$, kept in a bounded range & $J_p$ \\
\rowcolor{fgGrey} Transfer & PIC, no affine term in P2G & APIC & APIC with FLIP share $\chi$ \\
\bottomrule
\end{tabular}
\end{table}

\textbf{Fracturing solids.}
As in CD-MPM~\cite{wolper2019cdmpm}, the damage is driven by a history field $H_p=\max\big(0,\max_t(\psi^+-\psi_{\rm cr})\big)$, the largest excess so far of the tensile part $\psi^+$ of the strain energy density over a threshold $\psi_{\rm cr}$.
The field is updated only while the material dilates, so crumb confined around a pressing finger does not fracture.
The damage relaxes toward $H_p/(H_p+\alpha_d)$, with a softening constant $\alpha_d$ and a time constant that sets the brittleness.
The degradation $g(d_p)$ always scales the deviatoric stress and scales the volumetric stress only under dilation.
Because the yield stress falls with the damage, intact crumb dents only mildly, while material in the process zone draws into strands before it ruptures.

\textbf{Granular solids.}
Sintering is modeled by a stretch limit $\theta_s$ that stays small while the snow is loose and ramps to the value of Stomakhin \etal~\cite{stomakhin2013snow} once the compaction $1-J^P_p$ passes a threshold.

\textbf{Liquids.}
The liquid is weakly compressible, with an artificial bulk modulus $K$; the resulting sound speed sets the time step.
The pressure is set to zero for $J_p>1$ so that drops in flight do not pull on their neighbors, and the volume ratio is updated as $J_p\leftarrow J_p(1+h\,\mathrm{tr}\,\mathbf C_p)$, because the multiplicative update of $\mathbf F_p$ would inflate $J_p$ under rotation.

\textbf{Colliders.}
The projection $\Pi_i$ of \cref{eq:grid} enforces the collider conditions of \cref{sec:solid} at the grid nodes.
The spheres of a robot link advance with its velocity at every substep between control frames and carry one friction coefficient each; where the trajectory sets a grip, the nodes within a thin skin of a sphere take its velocity.
At the walls of tools and containers, the conditions hold at the nodes near a face, and any particle found inside a wall is pushed out.
The table is a frictional plane.

\section{Coupled Shell and Air Sub-Solver}
\label{sec:supp-air}
\textbf{Problem.}
The air fills the domain $\Omega_a$ outside the solids and the sheet; it is incompressible and inviscid, with density $\rho$, velocity $u$ and pressure $p$, and obeys
\begin{equation}
\rho\big(\partial_tu+(u\cdot\nabla)u\big)=-\nabla p,\qquad \nabla\!\cdot u=0\quad\text{in }\Omega_a,
\label{eq:euler}
\end{equation}
with $p=0$ on the open boundary of the domain and a prescribed normal velocity on solids such as the floor and the hair dryer.
The air reaches the sheet from both sides, where it may slide along the sheet but not cross it: $u\cdot n=\mathbf v\cdot n$ on either side, with $\mathbf v$ the velocity and $n$ the normal of the sheet.
In return, the sheet experiences the jump between the pressures $p^-$ and $p^+$ on its two sides as a traction $(p^--p^+)\,n$.
The sheet is the FEM shell of Appendix~\ref{sec:supp-contact}, contact and friction included, with this traction added.
A time step separates the advection of the air, which is explicit, from what must hold implicitly at the end of the step: incompressibility, the implicit Euler step of the shell with contact, and the two conditions on the sheet.

\textbf{Air grid.}
On the octree of Cirrus~\cite{wang2025cirrus}, each leaf cell $c$ holds one value of the pressure impulse $q=hp$, and each face $f$, with axis $e_f$, area $A_f$, and lower and upper cells $c_f^-$ and $c_f^+$ along $e_f$, holds the velocity component $u_f$ along $e_f$.
With $\delta_f$ the distance between the two cell centers and $V_f=A_f\delta_f$ the control volume of the face, the divergence $D$ and its transpose are
\begin{equation}
(Du)_c=\textstyle\sum_{f\in\partial c}s_{c,f}A_fu_f,\qquad (D^{\top}q)_f=A_f\big(q_{c_f^-}-q_{c_f^+}\big),
\label{eq:div}
\end{equation}
where the sum runs over the faces $\partial c$ of cell $c$, with $s_{c,f}=+1$ if $c=c_f^-$ and $-1$ if $c=c_f^+$, so $-D^{\top}q$ is the difference of $q$ across each face times its area.
Faces are open ($\mathcal O$), solid with a prescribed velocity $u_{\mathcal S}$ ($\mathcal S$), or walls ($\mathcal W$), and cells on the open boundary have $q=0$.
A subscript restricts $D$ or $u$ to one of these sets, and $\rho V$ and $\beta$ stand for the diagonal matrices of the face masses $\rho V_f$ and of the penalties $\beta_f$.
The coupled step uses only the result of the advection, the advected velocity $u^*$.
Without a sheet, the projection $\min_u\frac12\sum_f\rho V_f(u_f-u^*_f)^2$ subject to $Du=0$ gives $u=u^*+(\rho V)^{-1}D^{\top}q$ with $Lq=-Du^*$, which is solved by conjugate gradients with a multigrid preconditioner.

\textbf{Sheet as walls.}
Cells within a band around the sheet are kept at the finest level.
A wall $f$ is crossed by a sheet triangle $t_f$ at the barycentric weights $b_f$, and the air sees the normal speed of the sheet at the crossing, scaled by the tilt $\tau_f$,
\begin{equation}
u_f=\tau_f(N\mathbf v)_f,\qquad (N\mathbf v)_f=n_f^{\top}\textstyle\sum_{i\in t_f}b_{f,i}\mathbf v_i,
\label{eq:wallspeed}
\end{equation}
so tangential motion of the sheet creates no flux, and air flows around its open edges.
Walls, normals and tilts are frozen for the step.

\textbf{Coupled problem.}
For walls that move at the normal speeds $\nu$, the projection of \cref{eq:ea} has the minimizer $u_{\mathcal O}=u^*_{\mathcal O}+(\rho V)^{-1}D_{\mathcal O}^{\top}q$ with $Lq=-r(\nu)$, which gives the closed form of $e_a$ and its Hessian $\nabla^2e_a=C^{\top}L^{-1}C$.
The minimizer of \cref{eq:coupled} therefore satisfies
\begin{equation}
\begin{aligned}
\nabla e_s(\mathbf v)&=N^{\top}C^{\top}q,\\
Lq&=-\big(D_{\mathcal O}u^*+CN\mathbf v+D_{\mathcal S}u_{\mathcal S}\big).
\end{aligned}
\label{eq:coupled-kkt}
\end{equation}
These are the two optimality conditions named in \cref{sec:gas}, in that order; the contact multipliers of Appendix~\ref{sec:supp-contact} are left out of the first line for brevity.
On wall $f$, the load is $J_f=(C^{\top}q)_f=\tau_fA_f(q_{c_f^-}-q_{c_f^+})$, the jump of the pressure impulse across the sheet times the area of the wall seen along $n_f$, and $N^{\top}$ spreads it to the vertices of the triangle.

\textbf{Splitting.}
The energy $e_a$ couples all walls through $L^{-1}$, so its Hessian is never formed.
With $\nu$, $J$ and $\beta_f$ as in \cref{sec:gas}, the augmented Lagrangian
\begin{equation}
\mathcal L_\beta=e_s(\mathbf v)+e_a(\nu)-J^{\top}(N\mathbf v-\nu)+\textstyle\sum_f\frac{\beta_f}{2}\big((N\mathbf v)_f-\nu_f\big)^2
\label{eq:admm}
\end{equation}
is minimized alternately over $\mathbf v$ and $\nu$, and each such pair of minimizations is followed by the dual update $J\leftarrow J-\beta(N\mathbf v-\nu)$.
The minimization over $\mathbf v$ is \cref{eq:shellsweep}.
Over $\nu$, stationarity gives $\nu=\bar\nu+\beta^{-1}C^{\top}q$ with $\bar\nu=N\mathbf v-J/\beta$; substituting this into $Lq=-r(\nu)$ gives \cref{eq:robin}, after which the dual update reduces exactly to $J=C^{\top}q$ (\cref{eq:dual}).
In the Poisson solver, $C\beta^{-1}C^{\top}$ adds to each wall the face coefficient $\kappa_f=A_f^2\tau_f^2/\beta_f$, which takes the place of the coefficient $A_f^2/(\rho V_f)$ of an open face.
Algorithm~\ref{alg:aero} lists the~step.

\begin{algorithm}[t]
\caption{Coupled shell and air step. Every $\arg\min$ over $\mathbf v$ is the FEM step of Algorithm~\ref{alg:fem} with contact and friction.}
\label{alg:aero}
\small
\begin{algorithmic}[1]
\State Build the walls at $\mathbf x^n$, form $N$ and $C$, label the pockets
\State $u^*\gets$ advected $u^n$
\State $J\gets C^{\top}\tilde q^{\,n}$ \Comment{last commit's impulse}
\State $\mathbf v\gets\arg\min_{\mathbf v}\,e_s(\mathbf v)-J^{\top}N\mathbf v$ \Comment{sweep 0}
\For{$k=0,1,\dots,K_{\max}$}
\State $\bar\nu\gets N\mathbf v-J/\beta$
\State $\bar r\gets D_{\mathcal O}u^*+C\bar\nu+D_{\mathcal S}u_{\mathcal S}$
\State Solve $(L+C\beta^{-1}C^{\top})\,q=-\bar r$ \Comment{air}
\State $\nu'\gets\bar\nu+\beta^{-1}C^{\top}q$, \ $J\gets C^{\top}q$
\If{$k\ge1$ and $r_p,r_d\le\epsilon_a+\epsilon_r\Delta v$}
\State \textbf{break}
\EndIf
\State $\nu\gets\nu'$, \ $\hat\nu\gets\nu+J/\beta$
\State $\mathbf v\gets\arg\min_{\mathbf v}\,e_s+\sum_f\frac{\beta_f}{2}\big((N\mathbf v)_f-\hat\nu_f\big)^2$ \Comment{shell}
\EndFor
\If{the sheet fails the geometric check}
\State Discard both sides and redo the step with $h/2$
\EndIf
\State Commit the shell, $\mathbf v^{n+1}\gets\mathbf v$, $\mathbf x^{n+1}\gets\mathbf x^n+h\mathbf v$
\State $w\gets\tau\,N\mathbf v^{n+1}$, corrected by one constant per pocket
\State Solve $L\tilde q=-(D_{\mathcal O}u^*+D_{\mathcal W}w+D_{\mathcal S}u_{\mathcal S})$ \Comment{closed}
\State $u^{n+1}_{\mathcal O}\gets u^*_{\mathcal O}+(\rho V)^{-1}D_{\mathcal O}^{\top}\tilde q$, \ $u^{n+1}_{\mathcal W}\gets w$
\end{algorithmic}
\end{algorithm}

\textbf{Sweeps and convergence.}
Sweep~0 (lines~3 and~4) leaves out the wall term, because a cold start would let the penalty pull the first iterate halfway back to the previous velocity.
Sweep~0 alone is the lagged, explicitly staggered scheme, and sweep~0 with one regular sweep is a one-pass scheme with Robin transmission conditions~\cite{badia2008robin,fernandez2013robin}, which is stable at any added-mass ratio; further sweeps remove its splitting error.
For the linearized problem, with a fixed contact set, a quadratic shell energy and exact solves, a regular sweep contracts each mode by the factor
\begin{equation}
\varrho(\beta)=\frac{ab+\beta^2}{(a+\beta)(b+\beta)},
\label{eq:rho}
\end{equation}
where $a$ is the effective mass of the mode on the shell side, its mass plus $h^2$ times its stiffness, and $b$ is the mass of the air that moves with it.
With $\beta_f=m_f$, the penalties sum to the mass of the sheet at every orientation, so a rigidly moving patch has $a=\beta$ and $\varrho=\frac12$ for every $b$; for a normal motion pinned by contact, $a\to\infty$, the factor tends to $b/(b+\beta)$, which is the slower case when the moving air outweighs the sheet.
The sweeps stop (line~10) when the mismatch $r_p=\max_f|(N\mathbf v)_f-\nu'_f|$ between the wall speeds of the two sides and the change $r_d=\max_f|\nu'_f-\nu_f|$ of the wall speeds on the air side fall below an absolute tolerance $\epsilon_a$ plus a relative tolerance $\epsilon_r\Delta v$, or when the sweep count reaches its cap $K_{\max}$.
Here $\Delta v$ is the largest velocity change of a sheet vertex in the step, bounded below by $h|\mathbf g|$; a tolerance relative to a flow speed would instead stop the sweeps after the first whenever the sheet barely accelerates.
Anderson acceleration on $(\nu,J/\beta)$, not shown in Algorithm~\ref{alg:aero}, reduces the number of sweeps.
With contact, the shell problem is nonconvex and solved inexactly, so convergence is measured, not proven.

\textbf{Commit.}
The step is committed only if the sheet passes a geometric check after the last sweep: no vertex may have moved by more than the width of a cell at the finest level, beyond which the walls frozen at $\mathbf x^n$ can no longer stop flux through the sheet.
If the check fails, the shell and the air are both reset and the step is redone with $h/2$ (lines~16 to~18).
The commit (lines~19 to~22) is the closed-wall projection of \cref{sec:gas} at the committed wall velocity $w$, and its pressure impulse $\tilde q$ starts the next step.
With closed walls, the projection in a pocket $\mathcal P$, such as air trapped between folded layers, is solvable only if the net wall flux $Q_{\mathcal P}$ of the pocket vanishes.
The wall velocities of each pocket are therefore shifted by one constant, obtained from a small symmetric positive definite system that couples the pockets sharing a wall, and each pocket keeps the mean pressure of the last Robin solve.
The removed volume $h|Q_{\mathcal P}|$ is the leak that the step reports (\cref{sec:signals}).

\section{Schema Records and Harness}
\label{sec:supp-schema}
\Cref{sec:schema} defines a schema and the loop that uses it.
This appendix adds the content of the text that the agent starts from, the way observables become events, the format of a stored record, the five records formed from the rope rehearsals with their evidence and revision, the archive across tasks, the recall of a record on a new task, and the sub-agents of the harness.

\textbf{Seed text.}
The text that the agent starts from (\cref{sec:schema}) asks it to identify the relations of a situation that stay the same when the object, the viewpoint or the embodiment changes; to compare them with the cases it has already solved; to retrieve the closest stored schema; to adapt it to the current embodiment; and, when no stored schema matches well, to look for a structurally similar one in a more distant task before writing a new skill.

\textbf{From observables to events.}
For every saved sample of a rehearsed run and every body $a$ that touches the object (a robot hand or the table), the agent reads the channels of \cref{tab:channels}.
The first three sum the contact observables of \cref{sec:signals} over the contact pairs of $a$, for the whole object and for each of its numbered material regions, and the last measures the motion of the object.
An event is a window of consecutive samples of one run in which a channel changes state, for example a contact count that reaches zero, a vertical force that changes sign, or a load that moves from one support to another.
The agent proposes events from such changes in the logs and stores each under a hash identifier; later rounds may add events but do not rewrite an event that a schema cites.
An evidence item then cites an event, a channel, a range of samples and a statistic over that range (the first or last value, the minimum, the maximum or the change).
The events of \cref{fig:schema} are of this kind: the grip is lost when the contact count of a hand on its region reaches zero, and the load transfers when the vertical force of the table rises while that of a hand~falls.

\begin{table}[t]
\caption{\textbf{Channels} that the agent reads per saved sample, as logged for the rope; $a$ is a robot hand or the table, and $r$ is a material region of the object.}
\label{tab:channels}
\centering\footnotesize
\setlength{\tabcolsep}{4pt}
\setlength{\aboverulesep}{0pt}\setlength{\belowrulesep}{0pt}
\renewcommand{\arraystretch}{1.14}
\begin{tabular}{@{}l>{\raggedright\arraybackslash}p{4.25cm}l@{}}
\toprule
Channel & Meaning & Unit \\
\midrule
count($a$) & number of contact pairs between $a$ and the object & -- \\
\rowcolor{fgGrey} normal($a$) & total normal load of $a$ on the object & N \\
force$_z$($a$) & vertical force that $a$ applies to the object, positive upward & N \\
\rowcolor{fgGrey} count($a,r$), \ldots & the three channels above restricted to region $r$ & \\
speed & largest displacement of a centerline node between two saved samples, divided by their time gap & m/s \\
\bottomrule
\end{tabular}
\end{table}

\textbf{Record format.}
The agent stores each schema as one JSON record with the nine fields of \cref{tab:fields}; \cref{fig:schema} (right) shows one in abridged form.
The last column of \cref{tab:fields} maps the tuple $\sigma$ of \cref{sec:schema} onto the fields, in a form that keeps every statement tied to the samples behind it.
The failure signatures $F$, for example, are the evidence of the runs that failed, together with the signs that the guidance marks as insufficient, and the parameters $\Theta$ are the values from the runs that succeeded, stored with those runs.
The guidance states the order of the steps, which is the phase graph $G$ of the schema, and the assembly of a task keeps the phase order of the whole task.

\begin{table}[t]
\caption{\textbf{Fields of a schema record} and the part of $\sigma=\langle R,\Phi,P,G,F,\Theta\rangle$ that each one holds.}
\label{tab:fields}
\centering\footnotesize
\setlength{\tabcolsep}{4pt}
\setlength{\aboverulesep}{0pt}\setlength{\belowrulesep}{0pt}
\renewcommand{\arraystretch}{1.14}
\newcommand{\rr}{\raggedright\arraybackslash}
\begin{tabular}{@{}>{\rr}p{1.7cm}>{\rr}p{5.05cm}c@{}}
\toprule
Field & Content & $\sigma$ \\
\midrule
id, name & identifier and the experience in one line & \\
\rowcolor{fgGrey} status & candidate experience until the schema is confirmed on other objects & \\
physical context & the interaction, the roles with their binding in every run, and the conditions under which the schema applies & $R$ \\
\rowcolor{fgGrey} experience & the relation that was observed, marked as observed, with the evidence it rests on & $\Phi$ \\
use guidance & a purpose (observe, diagnose or consider an adjustment), what to do, and when this is not sufficient & $G,F$ \\
\rowcolor{fgGrey} evidence & items marked as support or boundary, each an observation with its numeric references (event, channel, samples, statistic, value) & $P,F,\Theta$ \\
boundaries & the scope that was observed, the transfer hypothesis and the unknowns & \\
\rowcolor{fgGrey} diagnostic requirements & the quantities needed to test the schema again, and why each one matters & $P$ \\
\bottomrule
\end{tabular}
\end{table}

\begin{table*}[t]
\caption{\textbf{The five schema records} formed during the rehearsals of rope knotting, condensed from the stored records. Runs: number of rehearsed runs in which the record binds its roles. Forces are those applied to the rope; samples index the saved states of a run.}
\label{tab:records}
\centering\footnotesize
\setlength{\tabcolsep}{4pt}
\setlength{\aboverulesep}{0pt}\setlength{\belowrulesep}{0pt}
\renewcommand{\arraystretch}{1.16}
\newcommand{\rr}{\raggedright\arraybackslash}
\begin{tabular}{@{}>{\rr}p{1.95cm}>{\rr}p{2.25cm}c>{\rr}p{7.35cm}>{\rr}p{4.05cm}@{}}
\toprule
Schema & Roles & Runs & Evidence in the observables & Guidance \\
\midrule
\schema{Directional Support} & contact hand; flexible object & 4 & A hand that pressed the rope with 20.3\,N gave it 0.07\,N of upward force. At the final release, the left hand pressed with 8.2\,N and pushed the rope down with 2.4\,N while the table carried 12.8\,N. A 4\,cm carry lost regions 51 and 52 at sample 29; after the lower fingers were set again, the same carry kept both for 4.4\,s. & Judge support by the signed vertical force on the object and by the other supports, not by the squeeze. Set the supporting contacts again before repeating a carry that lost its region. \\
\rowcolor{fgGrey} \schema{Contact Identity} & manipulating hand; flexible object with distinct regions & 2 & The right hand's contacts fell to 0 and returned to 8 ten samples later; at the end it had 3, none on the region it had held and one on a new region. In a feed, the left hand kept at least 10 contacts and ended with 31.3\,N of normal load, while its contact moved from region 12 to region 11. & When the task needs a given region held, check that region and the part of the hand on it. The return of total contact only triggers the check. \\
\schema{Support Redistribution} & releasing hand; other support hand; environment support; placed object & 5 & A shallow descent lost the left hand's grip while the table still touched the rope; a deeper one kept it but dragged the little finger on the table. With that finger corrected, the table's force rose from 5.30 to 10.85\,N and the right hand's fell from 4.23 to 1.13\,N; both hands then left, and the table carried at least 10.35\,N. & Change how the object lands, check that the load has transferred and that idle fingers can leave, then release step by step. A contact with the table or one frame of upward force is not sufficient. \\
\rowcolor{fgGrey} \schema{Hold Validation} & supporting hand; flexible object & 4 & Under a hold command, the contact on the held region fell to 0 while 5 other contacts remained and the rope moved at up to 1.99\,m/s. After the hand was centered again, at least 10 contacts stayed and the rope moved at 0.03\,m/s. A hold that began with 15.5\,N of normal load lost contact within 10 samples. & Insert a hold before a handover or a feed, and continue only if the region's contact persists and the rope's motion decays. A hold command is not a check. \\
\schema{Feed Boundary} & carrying hand; flexible object & 9 & From one state, a hold and a 5\,cm carry along the rope kept the left hand's contact; a 5\,cm sideways carry lost it at sample 33, or at sample 53 with a turn of the wrist, and a 15\,cm carry along the rope lost it at sample 116. From another state, a feed lost the right hand's contact where a variant with less rotation kept at least 6~contacts. & Compare whole plans from the same state. When a short carry works, test a longer stroke and the hold after it instead of scaling the motion. \\
\bottomrule
\end{tabular}
\end{table*}

\textbf{Schemas from the rope runs.}
\Cref{tab:records} gives the five records that the agent formed from its rehearsals of rope knotting; three of them appear in \cref{fig:schema}.
Each is a statement about contact that a contact count or a single force would miss.
The two that \cref{fig:schema} does not show state that a hand commanded to hold may still lose the rope, and that a carry that works over 5\,cm may fail over 15\,cm.

\textbf{Forming and revising records.}
The records come from two review rounds over the logged rehearsals of one rope.
In the second round, the main agent re-screened the runs that it had not selected, examined 28 candidate runs sample by sample, and added 11 logged events to the 14 of the first round, which it left unchanged.
It added one record and extended two of the four existing ones: \schema{Directional Support} gained the comparison in which the support is set again before a carry, and \schema{Feed Boundary} grew from the choice of a feeding plan to the direction and stroke of a carry.
Two events were kept without a schema: their runs exceeded a joint limit, and one overlapped with an existing record while the other lacked a task-level metric.
A record separates what was observed from what is hypothesized.
A boundary item of the evidence is a run that limits the statement: the hold that stayed stable after the hand was centered shows that losing contact during a hold is not inevitable.
The unknowns name the confounds that remain: setting the support again changed the finger geometry, the load and the waiting time together, so \schema{Directional Support} states that no contact angle or force threshold follows from this comparison.
All five records keep the status of candidate experience, because they come from simulated runs of a single rope, reviewed by the agent that had formed the earlier records.
Each record states its transfer to other objects as a hypothesis, to be tested when a later task recalls it.

\textbf{Archive across tasks.}
\Cref{tab:experience-inventory} counts the event windows and the candidate schemas that the archive of the harness holds for seven tasks; several windows may come from the same run.
The archive extends beyond the two review rounds above with further rope experiments, which is why its rope row exceeds the 25 events and five records described here.

\begin{table*}[t]
\caption{\textbf{Archived experience of seven tasks}: stored event windows and candidate schemas, with representative event topics and one example of schema guidance per task.}
\label{tab:experience-inventory}
\centering\footnotesize
\setlength{\tabcolsep}{4pt}
\setlength{\aboverulesep}{0pt}\setlength{\belowrulesep}{0pt}
\renewcommand{\arraystretch}{1.14}
\begin{tabular}{@{}lrr>{\raggedright\arraybackslash}p{.295\linewidth}>{\raggedright\arraybackslash}p{.326\linewidth}@{}}
\toprule
Task & Events & Schemas & Example event topics & Example schema guidance \\
\midrule
Knot the rope   & 27 & 6 & Grasp loss, rope feed, load transfer & Check that the surface carries the load before release. \\
\rowcolor{fgGrey} Fold the T-shirt & 12 & 4 & Empty closure, cloth transport, settling & Check that the cloth moves with the grasp. \\
Shuffle the cards  & 7  & 3 & Layer slip, opposing contact, packet pose & Check support at individual card layers. \\
\rowcolor{fgGrey} Cut and serve the bread & 9  & 5 & Knife loading, carry loss, serving & Check retention of both halves after cutting. \\
Use the pipette & 18 & 9 & Tip ejection, plunger stroke, pouring & Check fluid delivery separately from plunger stroke. \\
\rowcolor{fgGrey} Tie the shoelace  & 16 & 5 & Lace motion, friction, settling & Track fixed material points relative to the wrist. \\
Fold a paper box & 16 & 9 & Loaded lift, flap insertion, unloading & Check that the box remains closed after release. \\
\midrule
\textbf{Total} & \textbf{105} & \textbf{41} & & \\
\bottomrule
\end{tabular}
\end{table*}

\textbf{Recall on a new task.}
When a later task recalls a record, the new binding of its roles is stored with the record, as \cref{fig:schema} (right) shows for the card shuffle; there, the guidance applies to the last subgoal, in which the cards settle on the table and the hands withdraw.

\textbf{Harness.}
The main agent owns the plan and the schema records.
Of its three sub-agents, the reviewer checks each plan before it runs, the diagnostician reads the observables of a failed rollout and names the relation that broke, and the verifier checks each subgoal against rendered views and the schema predicates.
On the robot, only their input changes, from the observables of \simname{} to the 3D reconstruction of the workcell.

\section{Real-World Pipeline}
\label{sec:supp-real}
\Cref{sec:rehearse,sec:perform} summarize how the agent watches the demonstration and performs on the robot.
This appendix details both, with further measurements from the rope run of \cref{sec:perform}.

\textbf{Watching.}
The hands and the object are reconstructed in every frame of the demonstration, the hands with the reconstructors cited in \cref{sec:rehearse}, and registered to the workcell frame, which gives a 4D reference of hand and object~\mbox{motion}.

\textbf{From rehearsal to goals.}
The rehearsed rollout is cut into stages, and each stage gives a goal: which hand takes which point of the object, and what state the object must reach.
A grasp goal is a vertex of the object mesh, selected from the simulated contact as in the cloth example of \cref{sec:perform}.
The rope rollout, 130\,s long in 64 segments, gives four goal states: the ends laid across each other, the crossing under the fist, the point on the tail that the index finger hooks (the hook point), and the ends after the knot.
A subgoal can thus span more than one stage, as the first subgoal of the rope in \cref{fig:h2s2r-a} spans the first two.
The T-shirt rollout gives four grasp goals.
The hand path between two goals is kept as a template relative to its grasp.

\textbf{Perceiving and registering.}
Each camera masks the object by color and back-projects its depth into the workcell frame.
The points are binned into 1\,cm voxels, and a voxel is kept if at least two cameras agree on it.
For the rope, the crossing is the highest stack of voxels, since two strands stand about 6\,cm high and one about 3.5\,cm.
The rehearsed crossing is moved onto the measured one, and the whole plan moves with it; the distance between the two is the registration shift reported in \cref{sec:perform}.

\textbf{Fitting the hand.}
The plan is solved by inverse kinematics in a kinematic model of the robot, from which the real hand deviates by a few centimeters.
Before each contact, the model hand is posed at the measured joint angles and registered to the points of the real hand in the camera views.
The resulting offset predicts where the hand would land, and the path is shifted and the hand fitted again until the predicted landing point is on the target; the two fits of \cref{sec:perform} are iterations of this loop.
Trajectories are played back at 30\,Hz with joint speeds below 45$^\circ$/s, and playback stops if a joint deviates from its command by more than 10$^\circ$.

\textbf{Checking and retrying.}
For the rope lift (\cref{sec:perform}), the voxels are counted at the crossing, and after a failed check the crossing is put back before the hand closes again.
Every later stage starts from a new measurement of the object and a new fit of both hands; in the rope run of \cref{sec:perform}, the hook point on the tail lay 6.5\,cm from the rehearsed one, and the index finger was brought to within 0.4\,cm of it.
The whole run took 7\,min\,23\,s.

\newcommand{\figsupppose}{%
\begin{figure*}[t]
\centering
{\figfont
\begin{tikzpicture}[x=1cm, y=1cm,
  fr/.style={inner sep=0pt, outer sep=0pt, append after command={\pgfextra{\draw[fgInk, line width=0.3pt] (\tikzlastnode.south west) rectangle (\tikzlastnode.north east);}}},
  tbadge/.style={fill=white, draw=fgInk, line width=0.3pt, inner sep=1.1pt, text height=4.1pt, text depth=1.1pt, font=\fontsize{5.6}{6}\selectfont},
  shdr/.style={anchor=base, font=\fontsize{6.4}{7}\selectfont, text=fgInk},
]
\def\W{17.4}\def\lw{0.46}\def\fg{0.04}
\pgfmathsetmacro\xf{\lw+0.06}
\pgfmathsetmacro\fw{(\W-\xf-7*\fg)/8}
\pgfmathsetmacro\fh{\fw*9/16}
\useasboundingbox (0,0.16) rectangle (\W,{-0.30-2*\fh-\fg});
\foreach \k/\st in {0/initial state, 1/crossed, 2/grasp the crossing, 3/lift, 4/thread, 5/loose knot, 6/pull the end, 7/knot}
  \node[shdr] at ({\xf+\k*(\fw+\fg)+\fw/2},-0.16) {\st};
\foreach \r/\pre/\lab/\ta/\tb/\tc/\td/\te/\tf/\tg/\th in {
    0/p1/Pose 1/00:11/02:22/03:15/04:33/05:26/10:30/12:41/13:43,
    1/p2/Pose 2/00:11/02:22/03:15/04:33/05:26/08:29/11:06/13:46}{
  \pgfmathsetmacro\ry{-0.30-\r*(\fh+\fg)}
  \fill[fgPeach] (0,\ry) rectangle (\lw,\ry-\fh);
  \node[rotate=90, inner sep=0pt, font=\fontsize{6.8}{8}\selectfont\bfseries, text=fgOrange] at (\lw/2,\ry-\fh/2) {\lab};
  \foreach \k/\t in {0/\ta, 1/\tb, 2/\tc, 3/\td, 4/\te, 5/\tf, 6/\tg, 7/\th}{
    \node[fr, anchor=north west] (f) at ({\xf+\k*(\fw+\fg)},\ry) {\includegraphics[width=\fw cm]{figures/supp/pose/\pre_\k.jpg}};
    \node[tbadge, anchor=north east] at ($(f.north east)+(-0.04,-0.04)$) {{\fontsize{4.6}{5}\selectfont\faClock[regular]}\hspace{1.5pt}\t};
  }
}
\end{tikzpicture}}
\caption{\textbf{Ablation of the initial state.} Real-robot rollouts of the same rope-knotting policy from two initial configurations of the arms and the rope. Frames are matched by stage, and the badge on each frame gives its elapsed time (min:s).}
\label{fig:pose}
\end{figure*}}

\newcommand{\cmprows}[5]{%
  \foreach \r/\v/\na/\nb/\nc in {#5}{
    \pgfmathsetmacro\ry{#1-\r*(#4+\fg)}
    \ifnum\r=2 \fill[fgLav] (\lw+0.06,\ry) rectangle (\xf-0.04,\ry-#4); \else \fill[fgGrey] (\lw+0.06,\ry) rectangle (\xf-0.04,\ry-#4); \fi
    \node[anchor=north west, inner sep=0pt, text width=\vt cm, align=left, font=\fontsize{6.6}{7.4}\selectfont\bfseries, text=fgInk] at (\lw+0.14,\ry-0.10) {\na};
    \node[anchor=north west, inner sep=0pt, text width=\vt cm, align=left, font=\fontsize{5.6}{6.5}\selectfont, text=fgSub] at (\lw+0.14,\ry-0.42) {\nb};
    \node[anchor=south west, inner sep=0pt, text width=\vt cm, align=left, font=\fontsize{5.6}{6.5}\selectfont, text=fgInk] at (\lw+0.14,\ry-#4+0.10) {\nc};
    \foreach \k in {0,...,#3}
      \node[fr, anchor=north west] at ({\xf+\k*(#2+\fg)},\ry) {\includegraphics[width=#2 cm, height=#4 cm]{figures/supp/compare/\cmppre_\v\k.jpg}};
  }
}
\newcommand{\figsuppcompare}{%
\begin{figure*}[tp]
\centering
\includegraphics[width=17.4cm,height=.88\textheight,keepaspectratio]{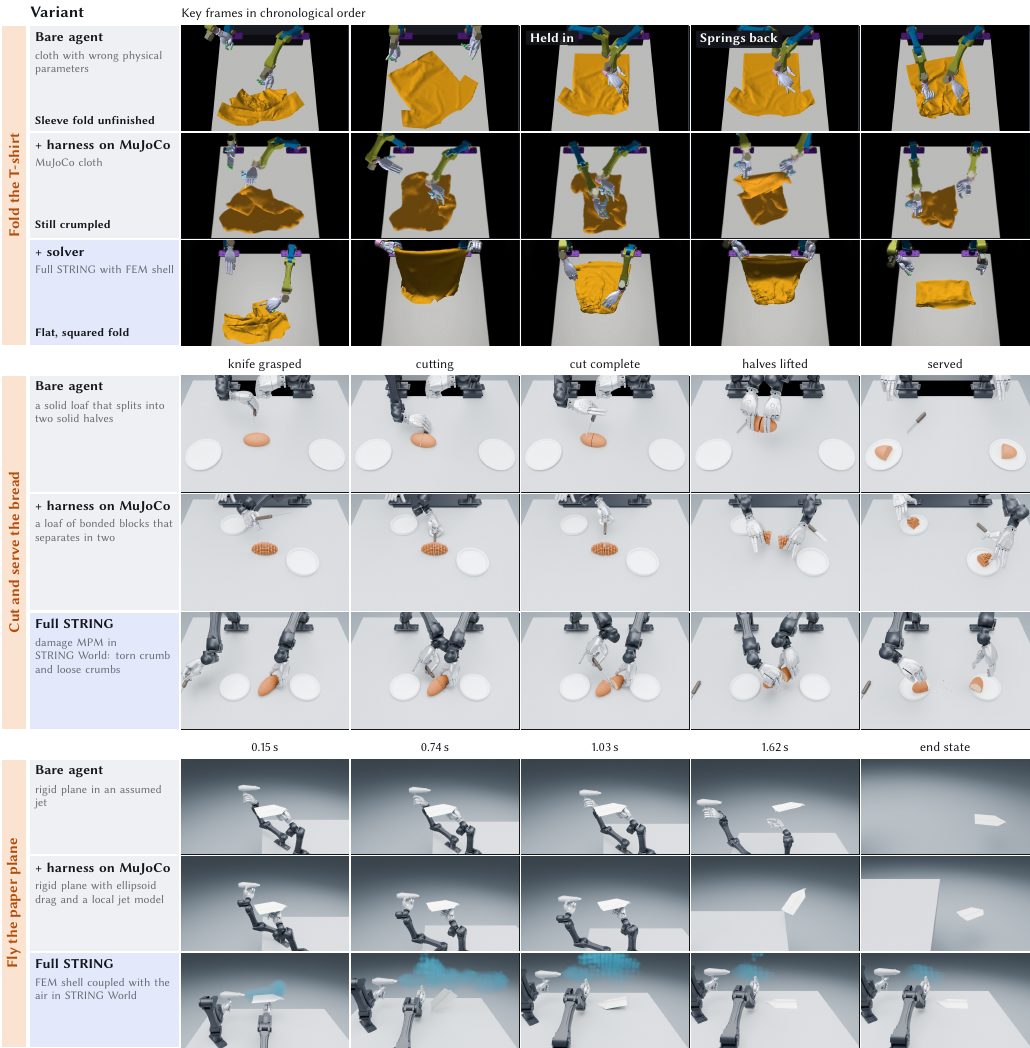}
\caption{\textbf{Ablation over time, in simulation, with the same agent.} Each block shows the agent without the harness or \simname{} (top row), with the harness on MuJoCo (middle) and with both (bottom, shaded). \textbf{T-shirt:} five key frames selected independently from each run in chronological order. In the bare-agent run, the sleeve is held inward at 62.4\,s (third frame) and springs back after release by 64.0\,s (fourth frame). \textbf{Bread:} frames matched by stage, since the three rollouts are not aligned in time. \textbf{Paper plane:} frames at the same simulated times from the start of each rollout; cyan shows the recorded air tracer in \simname{}. The line under each variant states how the run models the object, and \cref{tab:ablation-runs} gives the durations.}
\label{fig:baseline}
\end{figure*}}

\newcommand{\figsuppsimroll}{%
\begin{figure*}[t]
\centering
\includegraphics[width=17.4cm]{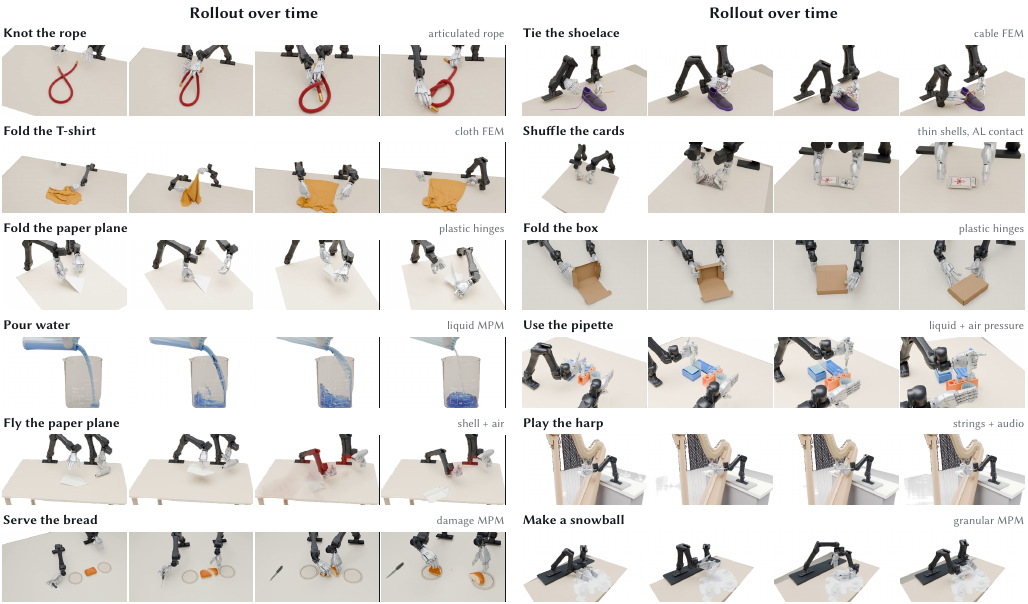}
\caption{\textbf{Rollouts in \simname{}.} The twelve simulated tasks of \cref{fig:multimodal} over time, four frames each from left to right, one row per group of materials: cables; cloth and cards; paper with plastic creases; liquids; air and audio; elastoplastic solids. Each sequence names its material model above the frames.}
\label{fig:simroll}
\end{figure*}}

\section{Rollouts in \simname{}}
\label{sec:supp-simroll}
\figsuppsimroll
\Cref{fig:simroll} extends the twelve task tiles of \cref{fig:multimodal} from one frame to four per task.
Ten of the twelve tasks are the real-robot tasks of \cref{tab:real}; pouring water and playing the harp are simulated only.

\section{Real-Robot Results on All Ten Tasks}
\label{sec:supp-results}\label{sec:supp-subgoal}
\newcommand{\hsrtask}[8]{%
  \pgfmathsetmacro\tp{-0.30-#1*(\bh+\bg)}
  \fill[fgPeach!55] (0,\tp) rectangle (\iw,\tp-\bh);
  \node[anchor=north west, inner sep=0pt, font=\fontsize{7.4}{8}\selectfont\bfseries, text=fgInk] at (0.09,\tp-0.10) {#3};
  \node[anchor=base west, inner sep=0pt, font=\fontsize{13}{13}\selectfont\bfseries, text=fgHi] (sr) at (0.09,\tp-0.80) {#4\kern0.5pt{\fontsize{7.5}{8}\selectfont\%}};
  \node[anchor=base east, inner sep=0pt, font=\fontsize{5.6}{6}\selectfont, text=fgSub] at (\iw-0.09,\tp-0.62) {average time};
  \node[anchor=base east, inner sep=0pt, font=\fontsize{6.6}{7}\selectfont\bfseries, text=fgInk] at (\iw-0.09,\tp-0.80) {#5};
  \draw[fgHi!30, line width=1.6pt] (0.09,\tp-0.92) -- (\iw-0.09,\tp-0.92);
  \draw[fgHi, line width=1.6pt] (0.09,\tp-0.92) -- ({0.09+(\iw-0.18)*#4/100},\tp-0.92);
  \node[anchor=north west, inner sep=0pt, text width=\tw cm, align=left, font=\fontsize{5.5}{6.3}\selectfont, text=fgSub] at (0.09,\tp-1.02) {#6};
  \foreach \sgn/\sgr/\sgm [count=\si from 0] in {#8}{
    \pgfmathsetmacro\sy{\tp-\bh+0.14+\si*0.262}
    \ifnum\sgm=1
      \node[anchor=base west, inner sep=0pt, font=\fontsize{5.6}{6}\selectfont\bfseries, text=fgInk] at (0.09,\sy) {\sgn};
      \node[anchor=base east, inner sep=0pt, font=\fontsize{5.6}{6}\selectfont\bfseries, text=fgHi] at (\iw-0.09,\sy) {\sgr};
    \else
      \node[anchor=base west, inner sep=0pt, font=\fontsize{5.6}{6}\selectfont, text=fgInk] at (0.09,\sy) {\sgn};
      \node[anchor=base east, inner sep=0pt, font=\fontsize{5.6}{6}\selectfont, text=fgInk] at (\iw-0.09,\sy) {\sgr};
    \fi
  }
  \node[anchor=base west, inner sep=0pt, font=\fontsize{5.2}{6}\selectfont, text=fgSub] at (0.09,{\tp-\bh+0.14+#7*0.262+0.03}) {subgoal success (\%)};
  \foreach \r/\kind/\lab/\col in {0/human/Human/fgLav, 1/sim/\simname{}/fgGrey, 2/real/Real robot/fgLav}{
    \pgfmathsetmacro\ry{\tp-\r*(\fh+\fg)}
    \fill[\col] (\iw+0.06,\ry) rectangle (\iw+0.06+\rw,\ry-\fh);
    \node[rotate=90, inner sep=0pt, font=\fontsize{5.2}{6}\selectfont\bfseries, text=fgInk] at (\iw+0.06+\rw/2,\ry-\fh/2) {\lab};
    \foreach \k in {0,...,6}
      \node[fr, anchor=north west] at ({\xf+\k*(\fw+\fg)},\ry) {\includegraphics[width=\fw cm]{figures/supp/h2s2r/#2_\kind\k.jpg}};
  }
}
\newcommand{\hsrpage}[1]{%
{\figfont
\begin{tikzpicture}[x=1cm, y=1cm,
  fr/.style={inner sep=0pt, outer sep=0pt, append after command={\pgfextra{\draw[fgInk, line width=0.3pt] (\tikzlastnode.south west) rectangle (\tikzlastnode.north east);}}},
  ghdr/.style={anchor=base, font=\fontsize{7.8}{8}\selectfont\bfseries, text=fgInk},
  shdr/.style={anchor=base, font=\fontsize{6.4}{7}\selectfont, text=fgSub},
]
\def\W{17.4}\def\iw{3.30}\def\rw{0.30}\def\fg{0.04}\def\bg{0.15}
\pgfmathsetmacro\tw{\iw-0.18}
\pgfmathsetmacro\xf{\iw+0.06+\rw+0.04}
\pgfmathsetmacro\fw{(\W-\xf-6*\fg)/7}
\pgfmathsetmacro\fh{\fw*9/16}
\pgfmathsetmacro\bh{3*\fh+2*\fg}
\useasboundingbox (0,0.14) rectangle (\W,{-0.30-5*\bh-4*\bg});
\node[ghdr, anchor=base west, inner sep=0pt] at (0.09,-0.16) {Task and success};
\node[shdr, anchor=base west, inner sep=0pt] at (\xf,-0.16) {\textbf{\color{fgInk}Stages of the task}\hspace{0.5em}each column shows the same stage in the three rows};
\draw[fgSub, line width=0.5pt, -{Stealth[length=3.4pt,width=3pt]}] (\W-1.55,-0.11) -- (\W-0.02,-0.11) node[pos=0, anchor=east, inner sep=2pt, font=\fontsize{6.4}{7}\selectfont, text=fgSub] {time};
#1
\end{tikzpicture}}}

\begin{figure*}[p]
\centering
\hsrpage{%
\hsrtask{0}{rope}{Knot the rope}{65}{13.6 min}{success: cross, thread through, knot}{3}{Grasp the rope end/70/0, Thread the loop/65/1, {Grasp, cross, grasp both strands}/70/0}
\hsrtask{1}{shoelace}{Tie the shoelace}{25}{15.8 min}{}{4}{Final pull/25/1, Second grasp/30/0, First grasp/35/0, Hold loops with both hands/35/0}
\hsrtask{2}{tshirt}{Fold the T-shirt}{60}{9.1 min}{success: sleeves folded in, folded in half}{3}{{Fold sleeves, fold in half}/60/1, Right-hand grasp/70/0, Grasp one corner/60/1}
\hsrtask{3}{cards}{Shuffle the cards}{80}{52 s}{success: no card flies out}{2}{Settle cards and withdraw/100/0, Riffle and release/80/1}
\hsrtask{4}{paper}{Fold a paper plane}{30}{10.3 min}{}{3}{Third fold/30/1, Second fold/30/1, First fold/60/0}
}
\caption{\textbf{From human demonstration to real robot, tasks 1--5.} Each block is one task: the single human demonstration (top row), the rehearsed rollout in \simname{} rendered in a scene matched to the lab (middle) and one real-robot rollout (bottom). The three clips are aligned in time, so every column shows the same stage of the task. Left: success rate over 20 trials, average execution time, the success criterion where one is stated, and the approximate success rate (\%) of each subgoal in order, with the hardest in orange.}
\label{fig:h2s2r-a}
\end{figure*}

\begin{figure*}[p]
\centering
\hsrpage{%
\hsrtask{0}{box}{Fold a paper box}{40}{3.7 min}{}{3}{Press closed/40/1, Fold side flaps inward/40/1, Lift and stand the box/70/0}
\hsrtask{1}{pipette}{Use the pipette}{40}{8.1 min}{}{5}{Eject tip/50/0, Transfer/55/0, Aspirate/40/1, Attach tip/50/0, Grasp/75/0}
\hsrtask{2}{bread}{Cut and serve the bread}{65}{4.0 min}{success: cut finished, served on the plates}{3}{Place halves on plates/65/1, Cut the bread/80/0, Grasp the knife/80/0}
\hsrtask{3}{snow}{Make a snowball}{65}{3.2 min}{}{3}{Set down the snowball/65/1, Compress with both hands/70/0, Gather the snow/80/0}
\hsrtask{4}{air}{Fly a plane by hair dryer}{30}{5.9 min}{}{3}{Blow the plane into flight/30/1, Pick up the plane/70/0, Pick up the hair dryer/40/0}
}
\caption{\textbf{From human demonstration to real robot, tasks 6--10}, in the layout of \cref{fig:h2s2r-a}. In the last block, the \simname{} row shows the air jet of the hair dryer in red.}
\label{fig:h2s2r-b}
\end{figure*}

\figsuppcompare\figsupppose %
\Cref{fig:h2s2r-a,fig:h2s2r-b} follow all ten tasks of \cref{tab:real} from the human demonstration through the rehearsal in \simname{} to a rollout on the robot, and \cref{tab:tasks} lists the object model and the hardest subgoal of each task.
The trials follow the protocol of \cref{sec:exp-real}; the subgoal rates are approximate, and in every task the overall rate equals the lowest of them.

\begin{table}[t]
\caption{\textbf{The ten real-robot tasks}: model of the object in \simname{} (\cref{fig:multimodal}; AL: augmented Lagrangian), hardest subgoal, success rate over 20 trials (\%) and average execution time.}
\label{tab:tasks}
\centering\footnotesize
\setlength{\tabcolsep}{3pt}
\setlength{\aboverulesep}{0pt}\setlength{\belowrulesep}{0pt}
\renewcommand{\arraystretch}{1.14}
\resizebox{\columnwidth}{!}{%
\begin{tabular}{@{}lllrr@{}}
\toprule
Task & Object model & Hardest subgoal & Succ. & Time \\
\midrule
Knot the rope & articulated rope & thread the loop & 65 & 13.6 min \\
\rowcolor{fgGrey} Tie the shoelace & cable FEM & final pull & 25 & 15.8 min \\
Fold the T-shirt & cloth FEM & corner grasp, final folds & 60 & 9.1 min \\
\rowcolor{fgGrey} Shuffle the cards & thin shells, AL contact & riffle and release & 80 & 52 s \\
Fold a paper plane & plastic hinges & second and third folds & 30 & 10.3 min \\
\rowcolor{fgGrey} Fold a paper box & plastic hinges & side flaps, closing & 40 & 3.7 min \\
Use the pipette & liquid + air pressure & aspirate & 40 & 8.1 min \\
\rowcolor{fgGrey} Cut and serve the bread & damage MPM & place the halves & 65 & 4.0 min \\
Make a snowball & granular MPM & set down the ball & 65 & 3.2 min \\
\rowcolor{fgGrey} Fly a plane by hair dryer & shell + air & blow into flight & 30 & 5.9 min \\
\bottomrule
\end{tabular}}
\end{table}

\textbf{Cables.}
The rope starts as an open arc on the table.
The hands cross its two ends into a loop, one hand closes on the crossing and lifts it so that both strands hang from the fist, the other hand threads an end through the loop, and the end is pulled until the knot closes.
The three subgoals lie within five percentage points of one another, at 70\%, 65\% and 70\%, and the hardest is threading the end through the loop.
The shoelace is the hardest of the ten tasks.
Both hands must hold loops of a thin lace, regrasp them twice and pull them apart to close the bow, and every subgoal is below 40\%: 35\% for holding the loops, 35\% and 30\% for the two grasps that follow, and 25\% for the final pull.

\textbf{Cloth and cards.}
The T-shirt starts crumpled.
In rehearsal and on the robot, one hand first lifts it by a corner until it hangs, the shirt is laid flat, the sleeves are folded in, and the body is folded in half.
The first grasp of a corner and the final folds each succeed in 60\% of the trials and set the overall rate, while the second grasp, by the right hand, succeeds in 70\%.
For the riffle shuffle, the hands hold the two halves of the deck on edge, let the cards fall alternately, release them into one pile and withdraw.
Once the halves are interleaved and released, the second subgoal, settling the deck and withdrawing the hands, has a rate of 100\%, so the failures of this task are cards that escape during the riffle.

\textbf{Paper with creases.}
Folding the paper plane and folding the box both depend on creases that stay where they are made.
The plane is folded on the table in three folds, of which the first succeeds in 60\% of the trials and the second and third in 30\% each; the later folds start from a sheet that the earlier ones have already creased.
The cardboard blank of the box is lifted and stood up, its side flaps are folded inward, and the box is pressed closed; standing it up succeeds in 70\% of the trials and each of the two later subgoals in 40\%.

\textbf{Liquids.}
Pipetting has five subgoals: grasping the pipette (75\%), attaching a tip (50\%), aspirating (40\%), transferring (55\%) and ejecting the tip (50\%).
Aspiration, in which the plunger is pressed and released with the tip in the liquid, is the bottleneck.

\textbf{Elastoplastic solids.}
The loaf is cut in two with a knife, and each half is carried to a plate.
Grasping the knife and cutting succeed in 80\% of the trials each, and the hardest subgoal is grasping the two halves and placing them on the plates (65\%).
For the snowball, loose snow in a tray is gathered with both hands, pressed between the palms and set down, with rates of 80\% for gathering the snow, 70\% for compressing it and 65\% for setting the ball down.

\textbf{Air.}
To fly the paper plane, one hand picks up the hair dryer (40\%) and the other the plane (70\%), and the jet then blows the plane off the hand (30\%).

\textbf{Execution time.}
The average time does not follow the number of subgoals: pipetting has five and takes 8.1\,min, whereas the rope has three and takes 13.6\,min.
The riffle is one continuous motion of both hands, while every stage of the other tasks adds the measurement, the fit and the possible retry of Appendix~\ref{sec:supp-real}.
The two slowest tasks are the two cable tasks, the shoelace and the rope, in which the cable is grasped anew for almost every subgoal.

\textbf{Across tasks.}
The rehearsed and the real rollouts pass through the same stages in the same order in all ten tasks.
In seven of them, the final subgoal is the hardest or shares the lowest rate; the exceptions are the rope, the cards and the pipette, where threading, the riffle and aspiration decide the~trial.

\section{Qualitative Ablation Results}
\label{sec:supp-ablation}
This appendix shows the frames behind the ablation of \cref{sec:exp-ablation}: the three variants on three simulated tasks, and the two initial states on the robot.

\subsection{Comparison with a General-Purpose Simulator}
\Cref{fig:baseline} follows over time the nine simulated runs whose outcomes \cref{sec:exp-ablation} summarizes, and \cref{tab:ablation-runs} gives the object model and the logged durations of the bread and plane runs.
The paragraphs below add what that summary omits.

\begin{table}[!t]
\caption{\textbf{Runs of the ablation} on the bread and the paper plane: simulated duration of the rollout and wall-clock time of the agent's session, as logged with each run (--: not logged). Neither was logged for the T-shirt runs.}
\label{tab:ablation-runs}
\centering\footnotesize
\setlength{\tabcolsep}{3pt}
\setlength{\aboverulesep}{0pt}\setlength{\belowrulesep}{0pt}
\renewcommand{\arraystretch}{1.14}
\newcommand{\rr}{\raggedright\arraybackslash}
\begin{tabular}{@{}l>{\rr}p{3.05cm}rr@{}}
\toprule
Variant & Model of the object & Sim. & Agent \\
\midrule
\multicolumn{4}{@{}l}{\emph{Cut and serve the bread}}\\
Bare agent & solid loaf, two solid halves & 117.2\,s & 4\,h\,06\,min \\
\rowcolor{fgGrey} + harness on MuJoCo & loaf of bonded blocks & 104\,s & 9\,h\,49\,min \\
Full \model{} & damage MPM & 134.5\,s & -- \\
\midrule
\multicolumn{4}{@{}l}{\emph{Fly the paper plane}}\\
Bare agent & rigid plane, assumed jet & 3.5\,s & 67\,min \\
\rowcolor{fgGrey} + harness on MuJoCo & rigid plane, ellipsoid drag and local jet & 3.2\,s & 69\,min \\
Full \model{} & FEM shell coupled with air & 2.46\,s & -- \\
\bottomrule
\end{tabular}
\end{table}

\textbf{T-shirt.}
The bare agent spreads the shirt and does not progress further.
In \simname{}, the harness first lifts the shirt and lays it flat, and then folds the sleeves and the~body.

\textbf{Bread.}
All three variants complete the task in their own model, so they differ in what the model can represent, not in whether the agent finishes.
In \simname{}, the crumbs fall on the table while the halves are carried to the plates.

\textbf{Paper plane.}
The log of the bare agent marks its assumed jet as a model that was not validated against a flight.
In \simname{}, the jet flows around the sheet while the hand holds it, and after release the plane glides down onto the table.

\textbf{Caveats and scope.}
The nine runs were recorded separately.
The \simname{} rollout of the T-shirt is smoothed for display, and that of the bread is joined from two segments, the second resumed from a checkpoint after the cut.
In the \simname{} run of the plane, the opening of the hand is scripted, the plane is pinned until release, and the release pose differs from that of the other two runs.
The comparison is therefore qualitative: it shows what each model can and cannot represent on the same task, and it does not measure a success rate.

\subsection{Different Initial States}
\Cref{fig:pose} shows the two real-robot runs behind the initial-state ablation of \cref{sec:exp-ablation}, which use the same policy with no new demonstration or rehearsal.
In the first, the rope lies in a narrow arc with one arm resting beside its end; in the second, it lies in a wider arc with both arms withdrawn.
Frames taken at the same elapsed times show the two runs at the same stage up to threading: the rope is crossed at 2\,min\,22\,s, lifted at 4\,min\,33\,s and being threaded at 5\,min\,26\,s.
After that, their timing differs: they end at 13\,min\,43\,s and 13\,min\,47\,s, and the second run ends one second after its last frame in \cref{fig:pose}.
That both runs pass through the same stages follows from the role binding noted in \cref{sec:exp-ablation}, which the harness carries out by moving the plan onto the measured rope before each stage (Appendix~\ref{sec:supp-real}): a different initial state changes where the first grasp lands, not the sequence of subgoals.

\end{document}